\documentclass{article}

    \PassOptionsToPackage{numbers, compress}{natbib}

\usepackage[preprint]{neurips_2023}

\usepackage[utf8]{inputenc} 
\usepackage[T1]{fontenc}    
\usepackage{hyperref}       
\usepackage{url}            
\usepackage{booktabs}       
\usepackage{amsfonts}       
\usepackage{nicefrac}       
\usepackage{microtype}      
\usepackage{enumitem}
\usepackage{graphicx}
\usepackage{multirow}
\usepackage{pifont}
\usepackage[dvipsnames]{xcolor}
\usepackage{tcolorbox}
\usepackage{adjustbox}
\usepackage{natbib}

\usepackage{cinzel}
\usepackage[T1]{fontenc} 
\newcommand{\modelname}{AssistGPT}
\newcommand{\aokvqa}{A-OKVQA}
\usepackage{mdframed}

\newcommand{\planner}{Planner}
\newcommand{\executor}{Executor}
\newcommand{\inspector}{Inspector}
\newcommand{\learner}{Learner}
\definecolor{mypurple}{RGB}{200,192,248}
\definecolor{mypurpledeep}{RGB}{142,126,240}
\definecolor{mygreen}{RGB}{117,170,156}
\definecolor{myyellow}{RGB}{255,192,0}
\definecolor{myblue}{RGB}{57,143,255}
\definecolor{mygrey}{RGB}{231,230,230}
\definecolor{codey}{RGB}{220,220,170}
\definecolor{coder}{RGB}{206,145,120}
\definecolor{codeb}{RGB}{156,220,254}
\definecolor{codenum}{RGB}{204,204,204}

\newcommand{\instruction}{{\small\colorbox{mygrey!100}{\textcolor{black}{\texttt{Instruction Prompt}}}}}
\newcommand{\inputq}{{\small\colorbox{myblue!50}{\textcolor{black}{\texttt{Input Query}}}}}
\newcommand{\observation}{{\small\colorbox{mygreen!50}{\textcolor{black}{\texttt{Observation}}}}}
\newcommand{\summary}{{\small\colorbox{mypurple!100}{\textcolor{black}{\texttt{Summary}}}}}
\newcommand{\summaryinput}{{\small\colorbox{mypurple!100}{\textcolor{black}{\texttt{Summary of Visual Inputs}}}}}

\newcommand{\dotp}{\ding{108} \textbf{\planner}}
\newcommand{\dote}{\textcolor{mygreen}{\ding{108}} \textbf{\executor}}
\newcommand{\doti}{\textcolor{mypurpledeep}{\ding{108}} \textbf{\inspector}}
\newcommand{\dotl}{\textcolor{myyellow}{\ding{108}} \textbf{\learner}}

\title{\includegraphics[scale=0.045, bb=-150 80 540 34]{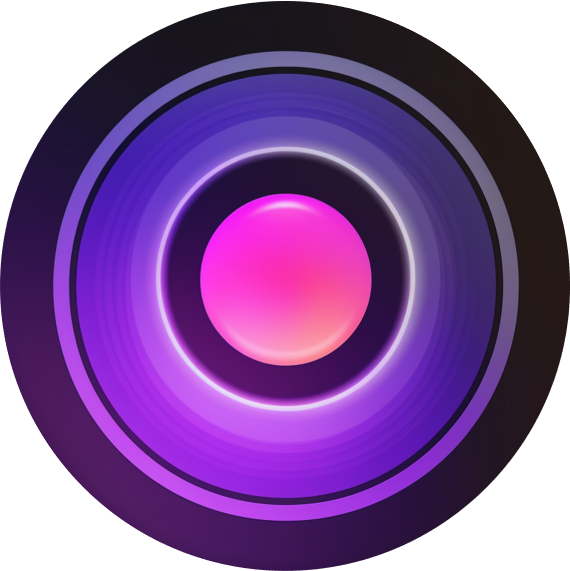}AssistGPT: A General Multi-modal Assistant that can Plan, Execute, Inspect, and Learn}

\definecolor{pearThree}{HTML}{E74C3C}
\definecolor{pearDark}{HTML}{2980B9}
\definecolor{pearDarker}{HTML}{1D2DEC}

\usepackage[dvipsnames]{xcolor}

\newcommand{\cmark}{\textcolor{mygreen}{\ding{51}}} 
\newcommand{\cxmark}{\textcolor{BurntOrange}{\ding{51}}} 
\newcommand{\xmark}{\textcolor{BrickRed}{\ding{55}}} 

\hypersetup{
 colorlinks,
 citecolor=pearDark,
 linkcolor=pearThree,
    breaklinks=true,
 urlcolor=pearDarker}
 
\author{Difei Gao,
    Lei Ji,
    Luowei Zhou, 
    Kevin Qinghong Lin, 
    Joya Chen,\\
    \textbf{
    Zihan Fan, 
    Mike Zheng Shou}\thanks{Corresponding author.} \\ 
    \\
    Show Lab, National University of Singapore, \\
    \url{https://showlab.github.io/assistgpt/}
}

\begin{document}

\maketitle

\begin{abstract}
Recent research on Large Language Models (LLMs) has led to remarkable advancements in general NLP AI assistants. Some studies have further explored the use of LLMs for planning and invoking models or APIs to address more general multi-modal user queries. Despite this progress, complex visual-based tasks still remain challenging due to the diverse nature of visual tasks. This diversity is reflected in two aspects: 1) Reasoning paths. For many real-life applications, it is hard to accurately decompose a query simply by examining the query itself. Planning based on the specific visual content and the results of each step is usually required. 2) Flexible inputs and intermediate results. Input forms could be flexible for in-the-wild cases, and involves not only a single image or video but a mixture of videos and images, e.g., a user-view image with some reference videos. Besides, a complex reasoning process will also generate diverse multimodal intermediate results, e.g., video narrations, segmented video clips, etc.
To address such general cases, we propose a multi-modal AI assistant, AssistGPT, with an interleaved code and language reasoning approach called Plan, Execute, Inspect, and Learn (PEIL) to integrate LLMs with various tools. Specifically, the Planner is capable of using natural language to plan which tool in Executor should do next based on the current reasoning progress. Inspector is an efficient memory manager to assist the Planner to feed proper visual information into a specific tool. Finally, since the entire reasoning process is complex and flexible, a Learner is designed to enable the model to autonomously explore and discover the optimal solution. We conducted experiments on A-OKVQA and NExT-QA benchmarks, achieving state-of-the-art results. Moreover, showcases demonstrate the ability of our system to handle questions far more complex than those found in the benchmarks.

\end{abstract}

\section{Introduction}
\begin{figure}[t]
  \centering
  \includegraphics[width=1\textwidth]{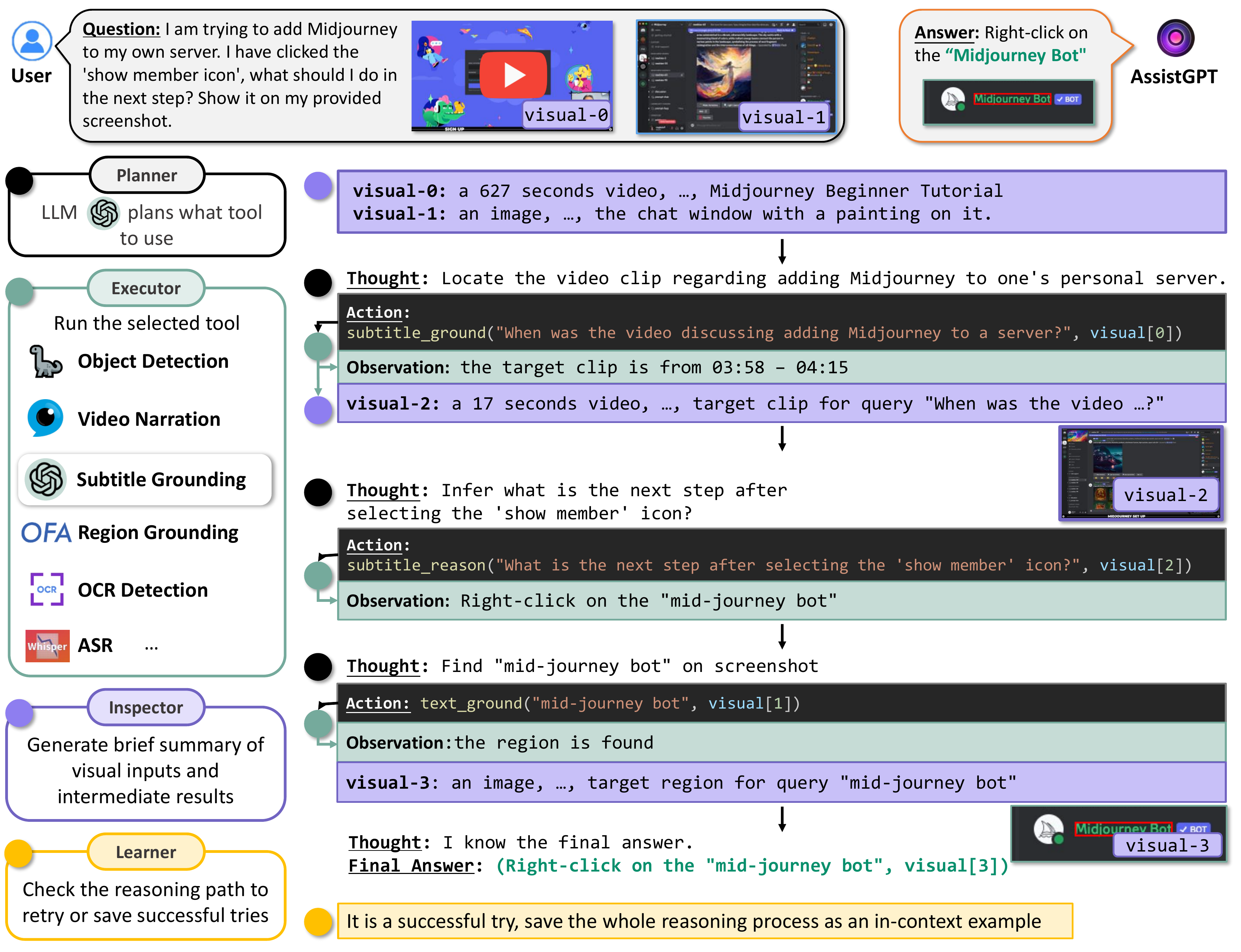}
  \caption{\textbf{In-the-wild example of AssistGPT.} AssistGPT can reason in an interleaved language and code format. Given a query input and visual inputs, AssistGPT plans the problem-solving path in language, using structured code to call upon various powerful tools. The Inspector, part of the system, can manage visual inputs and intermediate results, assisting the Planner to invoke tools. Meanwhile, the Learner can assess the reasoning process and collect in-context examples.}
  \label{fig1}
\end{figure}

Large language models (LLMs)~\cite{bert,gpt2,gpt3,llama}, especially ChatGPT~\cite{openai2022chatgpt}, have made remarkable progress in recent months, significantly advancing the field of developing AI assistants. Despite these advances, a single LLM serving as an AI assistant still exhibits inherent limitations in certain abilities, such as understanding visual environments and comprehending complex tasks, which restrict their utility in real-world applications. To address these shortcomings, a promising solution is to explore the integration and collaboration of multiple domain experts \textit{e.g.,} pretrained models or APIs, to tackle complex tasks. Numerous efforts have been made in this direction. Some works~\cite{zeng2022socratic, wang2022language, wu2023visual} utilize language as a bridge and transform the visual input into pure texts using foundational visual models, such as captioner~\cite{blip, blip2, dai2023instructblip}, object detectors~\cite{detr, Li_2022_CVPR, wu2022grit}, and OCR models~\cite{wang2022git, chen2022pali}. Subsequently, the extracted texts are fed into LLMs for reasoning tasks like question-answering.
Nonetheless, as for complex visual scenarios such as a long-form video with complicated scene switching, as shown in Fig.~\ref{fig1}, the generated texts may go well beyond the query requirements. This can lead to an abundance of superfluous information while crucial details relevant to the query may be omitted.

Some other concurrent works propose decomposing user queries into subtasks and plan to sequentially call external models or APIs to answer them. Currently, there are two branches of methods. The first one is language-based planning~\cite{shen2023hugginggpt,ge2023openagi,lu2023chameleon,li2023videochat}. For instance, HuggingGPT and Chameleon~\cite{shen2023hugginggpt, lu2023chameleon} propose using an LLM as a controller, managing and organizing the cooperation of expert models. 
Another branch of work is code-based planning~\cite{suris2023vipergpt, gupta2022visual,duan2023taskmatrix}. ViperGPT~\cite{suris2023vipergpt} proposes to use Codex to write the Python code to call visual-related APIs for handling multi-modal tasks. These approaches allow for invoking the models only when necessary, which allows models to only output useful information and optimize the use of computation resources. 

\begin{figure}[!ht]
  \includegraphics[width=\textwidth]{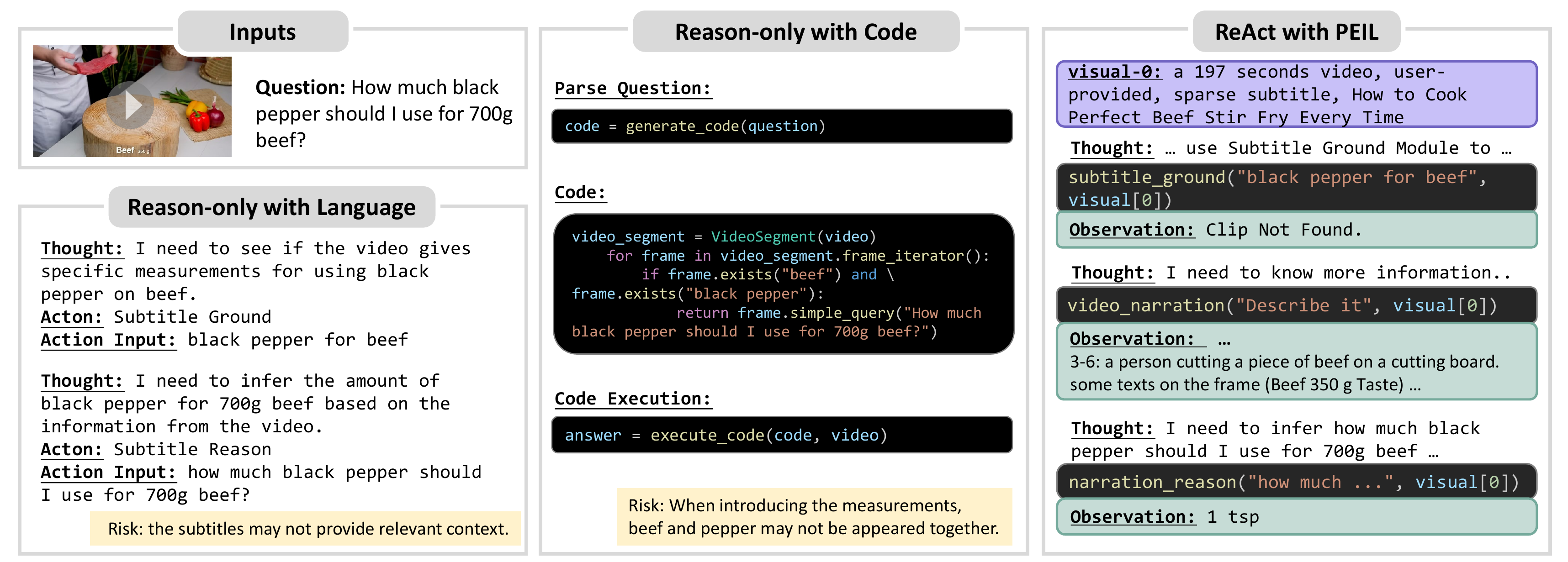}
  \caption{Comparison of PEIL and two mainstream reasoning methods in multi-modal tasks.}
  \label{fig2}
\end{figure}

Despite this progress, addressing high-level queries is still challenging. Specifically, current questions in existing benchmarks usually directly imply how to plan the reasoning. For example, for questions like "What is the red object used for?", no matter what the image is, the reasoning steps are relatively fixed, i.e., recognize the red object, then figure out its function. However, for more complex questions, there could be diverse reason paths. For example, for question \emph{``How much black pepper should I use for 700g beef?''} in Fig.~\ref{fig2}, the variations in the presentation of relevant information, whether it's in the form of subtitles, actions, text within videos, or a combination of these, can result in distinct reasoning paths. Therefore, as shown in Fig.~\ref{fig2}, once a reason-only approach makes a mistake, it becomes difficult for it to self-correct.

Similar approaches are already proposed in the NLP field, such as ReAct~\cite{yao2022react} and ToolFormer~\cite{schick2023toolformer}. However, there is a unique challenge in multimodal tasks: \emph{How to handle non-textual intermediate result?} For ReAct and ToolFormer, the outputs of external models can be directly fed into the Planner and passed to subsequent models. While the intermediate results obtained in multimodal tasks usually are cropped regions for the image grounding module, and segmented video clips for the temporal location module, as shown in Fig.~\ref{fig1}. In complex cases, it is hard for Planner to manage which information should be fed into the next module.

In this paper, we propose a multi-modal AI Assistant system, named AssistGPT\includegraphics[scale=0.03, bb=-150 80 540 34]{fig/assistgpt_logo_v2.png} (The design of our model's icon is inspired by the HAL 9000 from the movie ``A Space Odyssey'', a fictional artificial intelligence character), with interleaved language and code reasoning method, inheriting the advantages of flexible reasoning in ReAct and robust tool invocation in Program-based planning. Specifically, our system consists of four parts, Planner, Executor, Inspector, and Learner. We show how our system works in Fig.~\ref{fig1}. Similar to ReAct, the Planner thinks about what needs to be done next based on the current reasoning progress and invoking external models. What sets our method apart is the use of formatted code to invoke external models. The Executor wraps external tools into a uniform input and output format, allowing the tool to be invoked with structural commands. Simultaneously, we have also proposed an Inspector, which manages visual inputs and intermediate results during the reasoning process. It provides the Planner with summaries and metadata of all currently available visual materials. The combination of the Inspector and the Executor allows the model to efficiently implement complex reasoning. Moreover, it is challenging for the model to ensure correct reasoning in a zero-shot scenario. The Planner might output invalid code or unreasonable paths. To enable the system to continuously improve, we proposed the Learner, which checks whether the prediction process is reasonable or judges the correctness of the predicted results based on annotations. It allows the system to try multiple times and record successful examples as in-context examples.

The current version of \modelname~integrates 10+ tools for different functions, including image detection, captioning, region grounding, temporal grounding, OCR Module, object enumeration, speech-to-text, etc. By combining these functionalities, \modelname~can accomplish a wide range of multi-modal tasks which are still hard for existing systems. 

In summary, our contributions are as follows:
1) We constructed a general multimodal AI assistant that can accomplish diverse visual-related tasks with the cooperation of multiple models. 2) We propose a new compositional reasoning method that reasons over in an interleaved language and code manner. A simple learning mechanism is also proposed to improve the  \modelname's ability in planning.  3) We showcase \modelname's capabilities not only the benchmark results but also some realistic applications for processing complex images and long-form videos, understanding high-level queries, and handling flexible inputs.

\section{Related Work}
\textbf{Multi-modal Systems.}
Prior to the advent of LLM, remarkable works were done to design multimodal models for one or several specific tasks, such as focusing on visual appearance~\cite{antol2015vqa, anderson2018bottom, lu2019vilbert, tan2019lxmert, gao2020learning, gao2017visual}, visual-related knowledge~\cite{marino2019ok, aokvqa, wang2017fvqa, gui2021kat, marino2021krisp, gao2022cric}, action~\cite{bain2021frozen, Lei_2021_CVPR, wang2022geb+}, ego-centric videos~\cite{grauman2022ego4d, lin2022egocentric, gao2021env, hou2022cone}, instructional videos~\cite{wong2022assistq, lei2022assistsr, chen2023affordance}, scene text~\cite{hu2020iterative, yang2021tap, Gao_2020_CVPR, lei2022symbolic}, etc. They have achieved commendable results in specific tasks, however, their generalizability is relatively limited, making it challenging to address more complex and diverse questions in real-world scenarios.

Recently, two types of strategies are proposed for developing a general multi-modal system. 
One is pre-training LLM to support visual features as conditional inputs. The representative models are GPT-4~\cite{openai2023gpt4}, PaLM-E~\cite{driess2023palm}, BLIP-2~\cite{li2023blip}, and Mini-GPT4~\cite{zhu2022minigpt4}. Despite these methods being capable of directly processing multi-modal input, they still exhibit limitations in addressing advanced functional needs, such as image spatial grounding, long-form video grounding, and audio comprehension. Additionally, the computational cost of scaling these models can be extremely high. 
The alternative strategy aims to combine multiple models or APIs to accomplish complex multi-modal reasoning. For instance, models like the Socratic model~\cite{zeng2022socratic} and Visual ChatGPT~\cite{wu2023visual} achieve this by connecting ChatGPT with image generation models. HuggingGPT~\cite{shen2023hugginggpt} combines a variety of Huggingface models with LLMs. ViperGPT~\cite{suris2023vipergpt} employs Codex~\cite{chen2021evaluating} to call visual APIs via Python programming. 
Our \modelname~falls into the second category by combining and invoking various modules for multi-modal reasoning, but we propose a new framework PEIL for integrating external tools and models.

\textbf{Compositional Reasoning.}
Compositional reasoning methods in the field of visual question answering usually decompose questions into several subtasks, each addressed by a specific module. 
This kind of method offers strong interpretability due to its modular structure and the clear division of responsibilities among the individual components. 
This idea was initially put forward by ~\cite{andreas2016neural}. Subsequently, ~\cite{hu2017learning, johnson2017inferring} introduced an end-to-end variant based on LSTM and CNN. Traditional compositional reasoning methods are limited by language models' parsing capabilities, often requiring ground-truth question decomposition or reinforcement learning for optimal module usage.

With the advent of LLMs, question decomposition can be accomplished remarkably well in a zero-shot manner. Chain-of-thought prompts~\cite{wei2022chain}, Toolformer~\cite{schick2023toolformer}, and ReAct~\cite{yao2022react} enable models to plan how to solve an NLP problem. HuggingGPT~\cite{shen2023hugginggpt} and ViperGPT~\cite{suris2023vipergpt} are multi-modal systems that use LLM to parse a question into a series of reasoning steps. However, for complex queries, the model needs to determine the subsequent steps based on not only questions but also visual inputs or feedback from previously executed modules. MMReAct~\cite{yang2023mm} introduced the idea of ReAct to a multi-modal system to overcome it, while it is still under development and hasn't demonstrated its effectiveness on the benchmark.
Previous methods reason over either language reasoning or code, and as stated in the introduction, both have certain shortcomings.
Our work first proposes an interleaved language and code reasoning manner which can better handle general queries and complex visual inputs.

\textbf{Learning Schemes for Modular System.}
Early modular models primarily employed end-to-end Reinforcement Learning~(RL) to train each module's planning and acting from scratch. While this approach is practical for lightweight models, RL can introduce substantial overhead for systems where each module is an LLM. Toolformer~\cite{schick2023toolformer} proposes a self-supervised technique that optimizes planning requiring only a handful of demonstrations for each API. Specifically, Toolformer attempts various APIs to find successful examples and then fine-tunes the model. In contrast, we propose a straightforward mechanism in the multi-modal field, which can guide the system to retry and preserve the successful explorations as in-context examples.

\section{\modelname}

\begin{figure}[!ht]
  \centering
  \includegraphics[width=\textwidth]{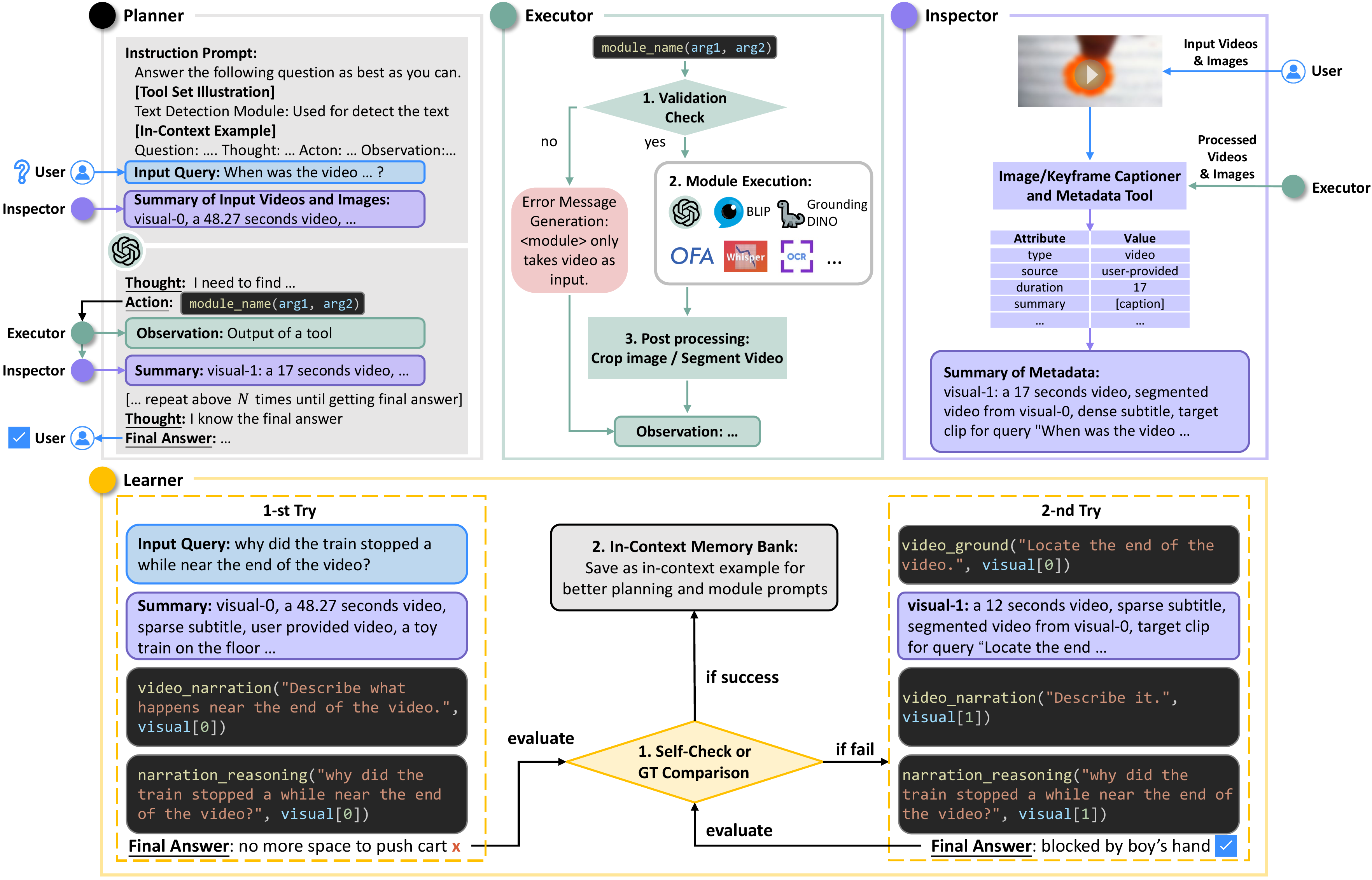}
  \caption{\textbf{Diagrammatic illustration of \modelname~system.} It consists of four core modules:\\
  \dotp: control the whole reasoning process;
  \dote: execute external tool and return feedback to \planner;
  \doti: manage the input and intermediate outcomes; 
  \dotl: assess the system performance and record successful trials as in-context examples.}
  \label{fig:overview}
\end{figure}

\textbf{Overview.}
\modelname~is a general multi-modal AI assistant system that can dynamically engage various tools in an interleaved language and code manner.
Specifically, given a general language query and reference images or videos as inputs, the goal of \modelname~is to generate the desired answer.
As shown in Fig.~\ref{fig:overview}, \modelname~is achieved by cooperation with four  core modules: \textbf{(a)} \planner, \textbf{(b)} \executor, \textbf{(c)} \inspector, and \textbf{(d)} \learner. 
The \planner~$\S$~\ref{planner} aims to control the whole reasoning process, with the \executor~$\S$~\ref{executor} supplying valuable feedback to \planner~by executing external tools.
The \inspector~$\S$~\ref{inspector} manages the input and intermediate results and assists the~\planner~in feeding proper content to the \executor. 
The \learner~$\S$~\ref{learner} is capable of assessing the system performance and record successful explorations as in-context examples. 
In the following sections, we will go through each module in detail.

\subsection{\planner}
\label{planner}
The \dotp~employs a highly intelligent LLM \textit{i.e.,} GPT-4~\cite{openai2023gpt4} as the central brain to control the global reasoning planning.
It begins the planning process by taking inputs from three types of information: an \instruction~consisting of the {\small\texttt{[Tool Set Illustration]}} and {\small{\texttt{[In-Context Example]}}}\footnote{The successful trials recorded by \learner, will be introduced later.}, \inputq, and the \summaryinput~created by \doti.

Then it generates the appropriate output for the next step, which consist of two parts: 
{\small\underline{{\texttt{Thought}}}}: a language phrase indicates what should be done next. While it doesn't affect the module or API call directly, it aids the LLM planning procedure. 
{\small\underline{{\texttt{Action}}}}: a structural string obeys the pre-defined template provided in the instructions. It specifies which external tool to call and what arguments to input, \textit{e.g.,}
{\scriptsize\colorbox{black}{\texttt{\textcolor{codey}{caption(}\textcolor{coder}{"what color is the car?"}\textcolor{codeb}{, visual}\textcolor{codey}{[}\textcolor{codenum}{0}\textcolor{codey}{])}}}}.

After each time \dote~call to an external tool, the tool returns outputs in the form of natural language, which we refer to as \observation. If the tool generates an intermediate outcome, \textit{e.g.,} a segmented video, our \doti~will store it and generate a \summary~for it. Both the \observation~and \summary~will be fed to the \dotp~to guide the planning of the next step. The following sections will introduce more details of {\small\underline{{\texttt{Action}}}}, \observation, and \summary.

Currently, we integrate 13 functional tools in \modelname~to power multi-modal assistance, as shown in Tab.~\ref{tab:tool}. These modules can be mainly categorized into three types: 
\begin{itemize}[leftmargin=0.2in, topsep=0pt, itemsep=0pt]

\item \textbf{Descriptor}: 
To effectively comprehend and utilize the data derived from intricate multimodal environments \textit{e.g.,} image, video, audio, and text, we employ a variety of fundamental models as basic descriptors for perception. These models, including (a) Image Caption, (b) Video Narration, (c) Object Detection, (d) Text Detection, and (e) ASR Translation, enable us to extract enough information from diverse source, thus enhancing our understanding of the multimodal sceanrio.
\item \textbf{Locator}: 
As the saying goes, a picture is worth a thousand words. Images or videos typically contain a wealth of information - objects, attributes, actions, events, and so on. However, the abundance of information can sometimes obstruct our problem-solving process. One crucial solution is to pinpoint the most crucial and valuable information from the rich sea of visual, textual, and audio data
This part incorporates several modules such as the (f) Region Ground, (g) Narration Ground, (h) Text Ground, and (i) Subtitle Ground.
\item \textbf{Reasoner}: 
The initial two sets of tools primarily deal with collect and identify of data, whereas the third set focuses on reasoning, utilizing the extracted information and external knowledge. This part incorporates modules such as (j) Knowledge Reason, (k) Narration Reason, (l) Subtitle reason, and (m) Temporal Reason modules. These modules primarily utilize LLM at their core by taking different types of information and prompts as inputs or a simple program.
\end{itemize}

\begin{table}[]
\centering
\caption{Module used in \modelname. A module may have different models, separated by a slash (/).}
\label{tool-table}
\resizebox{\textwidth}{!}{%
\begin{tabular}{lllllllll}
\cmidrule(r){1-4} \cmidrule(l){6-9}
Module Usage      & Core Model     & Input  & Output &  & Module Usage            & Core Model        & Input              & Output       \\ \cmidrule(r){1-4} \cmidrule(l){6-9} 
(a) Image Caption          & BLIP series~\cite{blip, blip2, dai2023instructblip}    & T, I & T 
&& 
(h) Text Ground      & Program + SSA~\cite{kirillov2023segment,chen2023semantic}  & T, I & T, I   
\\
(b) Video Narration        & BLIP series~\cite{blip, blip2, dai2023instructblip}       & T, V  & T  
&& 
(i) Subtitle Ground        & GPT~\cite{openai2022chatgpt}                & T, Sub.       & T, V         
\\
(c) Object Detection & G. Dino~\cite{liu2023grounding}~/~GLIP~\cite{Li_2022_CVPR}   & T, I & T      
&& 
(j) Knowledge Reason & GPT~\cite{openai2022chatgpt}            & T      & T      
\\
(d) Text Detection & Google OCR  & I    & T 
&& 
(k) Narration Reason       & GPT~\cite{openai2022chatgpt}                & T, Nar.       & T            
\\
(e) ASR Translation              & Whisper~\cite{radford2022robust}        & A  & T 
&&
(l) Subtitle Reason        & GPT~\cite{openai2022chatgpt}                & T, Sub.        & T           
\\
(f) Region Ground     & OFA~\cite{wang2022ofa}           & T, I & T, I  
&& 
(m) Temporal Reason       & Rule-based           & T, V               & T, V         
\\ \cmidrule(l){6-9} 
(g) Narration Ground       & GPT~/~CLIP~\cite{radford2021learning}          & T, Nar.       & T, V      & &
\multicolumn{4}{l}{\textbf{I}: Image, \textbf{V}: Video,  \textbf{T}: Text,  \textbf{A}: Audio, \textbf{Nar.}: Narration, \textbf{Sub.}: Subtitle}\\ \cmidrule(r){1-4} \cmidrule(l){6-9} 
\end{tabular}%
}
\label{tab:tool}
\end{table}

\subsection{Executor}
\label{executor}
The \dote~takes the code generated by the \dotp~as input, then call a module to produce the output by carrying out three steps to obtain the final result. These steps include validation check, module execution, and post-processing, as shown in Fig.~\ref{fig:overview}.

\begin{itemize}[leftmargin=0.2in, topsep=0pt, itemsep=0pt]
\item \textbf{Validation Check}: Even powerful LLM like GPT-4 can sometimes generate illegal code. For example, an image caption module accept a long video as input. We have designed a legality check for each module to determine whether the code is executable. Moreover, if code includes errors, we do not interrupt the entire reasoning process. Instead, we return an error message as the output code to the \dotp, allowing it to optimize the planning process in real-time.

\item \textbf{Module Execution}: We standard various modules or APIs into a unified interface using the code-style template \textit{i.e.,} \texttt{[Module\_Name](<text\_query>, <visual\_index>)}. Each module is designed to accept multiple text queries and visual data (images or videos) as input. In each standarded module, we  provide instructions on its function and the requirements of the argument, which is used for {\small\texttt{[Tool Set Illustration]}} in \dotp. Additionally, for the sake of simplicity and accuracy in Planning, the generated code is simplified. Later, a simple rule-based function will map it to the executable codes and then execute it to obtain the final result. 
 
\item \textbf{Post-processing}: For all modules, the generated results will be translated into a language format to inform the \dotp~about the outcome, as the \observation~part illustrated above. For instance, for the Narration Ground module, the model will return whether it has found the relevant segment. If so, output the start and end times of the segment. Additionally, many Ground-related modules will send their segmented video or cropped image region to the subsequent visual outcome manager \textit{i.e.,} \doti.

\end{itemize}

\subsection{Inspector}
\label{inspector}
The objective of the \doti~is to manage the visual inputs provided by the user and the intermediary results produced by our system to assist the \dotp~in deciding which source should be directed to which module. Specifically, the \doti~records the metadata of each visual element, which includes its type (image or video), source (provided by the user or generated by the system), and a brief description of the content (obtained from the caption model, or the title of an online video). 
For videos, there is some additional metadata, such as the duration of the video, whether it contains audio and subtitles. The \doti~monitors the inputs from the user and the outputs from the \dote. As soon as a new visual element is received, it appends the metadata, noted as \summary~in the above, to the reasoning history of the \dotp. With the cooperation of the \dotp, \dote, and \doti, our system can generate answers to difficult queries with complex visual inputs.

\subsection{Learner}
\label{learner}
Despite the robot generalization capabilities of LLMs, they can still easily encounter errors when dealing with multi-modal type queries. Thus, it is essential for an AI assistant to have self-valuate mechanism. 
To achieve this goal, we hope that the model can self-check the reasonableness of its output. On the other hand, when ground truth is available, we intend to gather successful prediction instances as in-context examples. Specifically, AssistGPT will repeatedly attempt to provide the answer when the response is not satisfactory until either passes the self-check, or correct answer is given (when ground truth is available) or a predefined maximum number of attempts is reached. \dotl~includes an evaluator implemented by the LLM, which operates in two modes: self-assessment and ground-truth comparison. These modes are activated depending on the availability of ground truth, and we discuss the two of them separately. 
\begin{itemize}[leftmargin=0.2in, topsep=0pt, itemsep=0pt]
    \item \textbf{Self-assessment mode} is activated when there is no user feedback or ground truth available. It takes the reasoning trace and the results of each step as input, allowing GPT to assess whether the reasoning is complete, consistent, and adhere to the required format. 

    \item \textbf{Ground-truth comparison mode} is activated when annotators provide ground truth. In this mode, GPT evaluates whether the AssistGPT's prediction is semantically consistent with the provided ground truth.
\end{itemize}

Furthermore, \dotl~encourages to keep trying until it receives positive feedback or reachs the maximum number of attempts. After conducting $N$ times explorations, several outcomes may arise: 

\begin{itemize}[leftmargin=0.2in, topsep=0pt, itemsep=0pt]
    \item \textbf{No adjustments required}: 
    If the model delivers the correct answer on its initial attempt, this suggests that AssistGPT can well-solve the current question effectively. Therefore, no improvement is required.
    \item \textbf{Plan Revision}: 
    If the model produces the correct answer after making $n$ attempts, where $1<n\leq N$, this implies that there is room for improving the model's planning capabilities. Therefore, we save the successful reasoning trace to {\small\texttt{[In-Context Memory Bank]}}. Consequently, when the model comes across a similar query in the future, it can use this as an in-context example.

    \item \textbf{Function Updates}: If the model still fails to provide the correct answer even after $N$ attempts, it is highly probable that the problem resides in a specific module or API rather than the planning process. It may necessitate incremental updates to the module. We will leave this for future work.
    
\end{itemize}

\section {Experiments}
\subsection{Experimental Setting}
\textbf{Datasets.}
Our system is evaluated on A-OKVQA~\cite{aokvqa} and NExT-QA~\cite{xiao2021next} benchmarks designed to test comprehensive multimodal capabilities, including visual facts, commonsense, temporal sequences, causality, etc. 
\textbf{\aokvqa}~\cite{aokvqa} is an innovative benchmark for knowledge-aware visual question answering with 25K questions that demand a high-level comprehension of commonsense and world knowledge. These questions in \aokvqa~go beyond the information contained in the image and cannot be answered solely by querying a knowledge base. 
Besides, the question is diverse, spanning a wide range of domains such as commonsense reasoning, visually-grounded, knowledge-based, and physical understanding. 
In our experiments, we assess the model performance under the in-context learning setting on the validation set, which consists of 1,145 questions. 
\textbf{NExT-QA}~\cite{xiao2021next} is a benchmark for evaluating the AI system's causal reasoning, temporal action reasoning, and rich object interactions in video question answering. NExT-QA has a total of 5,440 videos, with averaging 44 seconds in length, and approximately 52K manually annotated question-answer pairs. In our experiments, we assess the model performance under the in-context learning setting on the validation set, which consists of 4,996 questions.

\textbf{Implementation Details.}
In the following experiments, we use GPT-4 API provided by OpenAI~\cite{openai2023gpt4} as Planner. In the A-OKVQA experiments, we set Caption Module as BLIP2 or InstructBLIP (abbreviated as Ins.BLIP), use the Gounding Dino for the Object Detection model, and Google OCR for Text Detection. For the NExT-QA experiments, our Video Narration Module is based on InstructBLIP Vicuna-7B~\cite{dai2023instructblip}. Our experiments are performed on 4 A5000 GPUs.

\subsection{Quantitative Results}
\begin{table}[h]

\begin{minipage}{.55\linewidth}
\centering

\caption{Comparison of SOTAs on A-OKVQA dataset. D.A. and M.C. indicate direct answer and multi-choice. ICL: In-context Learning. ZS: Zero-shot inference.}
\resizebox{\textwidth}{!}{%
\begin{tabular}{llcc}
\toprule
                           & Model                                   & D. A. & M.C. \\ \midrule
\multirow{4}{*}{\rotatebox{90}{\textcolor{gray!40}{Sup.}}}  
                           & \textcolor{gray!40}{LXMERT~\cite{tan2019lxmert}}        & \textcolor{gray!40}{30.7}          & \textcolor{gray!40}{51.4}         \\
                           & \textcolor{gray!40}{KRISP~\cite{marino2021krisp}}         & \textcolor{gray!40}{33.7}          & \textcolor{gray!40}{51.9}         \\
                            
                           & \textcolor{gray!40}{GPV-2~\cite{kamath2022webly}}         & \textcolor{gray!40}{48.6}          & \textcolor{gray!40}{60.3}         \\
                           & \textcolor{gray!40}{InstructBLIP Vicuna-7B~\cite{dai2023instructblip}}   & \textcolor{gray!40}{64.0}  & \textcolor{gray!40}{75.7}         \\ \midrule
\multirow{3}{*}{\rotatebox{90}{ICL}} 
                           & PromptCap  ~\cite{hu2022promptcap}                  & \textbf{56.3}  & 73.2   \\
                           & AssistGPT (BLIP2 FlanT5$_{XL}$~\cite{li2023blip})   & 42.6          & 73.7   \\
                           & AssistGPT (Ins.BLIP Vicuna-7B)     & 44.3          & \textbf{74.7}   \\ \bottomrule
\end{tabular}%
\label{aokvqa-sota}
}

\end{minipage}%
\hspace{.02\linewidth}
\begin{minipage}{.42\linewidth}
\centering
\caption{Ablation study of our AssistGPT on A-OKVQA dataset. Ins.BLIP used here is the pre-trained version.}
\vspace{0.22cm}
\resizebox{0.95\textwidth}{!}{%
\begin{tabular}{clll}
\toprule
LLM                    & Model                 & D.A. & M.C.  \\ \midrule
\multicolumn{1}{c}{-}  & Ins.BLIP              & 13.4 & 53.8  \\ \midrule
                       & Ins.BLIP + GPT-4      & 27.9 & 55.2  \\
\multirow{5}{*}{\rotatebox{90}{AssistGPT}}  & Reason only           & 28.8 & 65.9  \\
                       & ReAct                 & 30.1 & 68.2  \\
                       & PIE                   & 32.4 & 72.4  \\
                       & PEIL w. Self-Check    & 41.2 & 74.2  \\ 
                       & PEIL w. GT-Check      & 44.3 & 74.7  \\  \bottomrule
\end{tabular}%
\label{aokvqa-ablation}
}
\end{minipage}
\vspace{-0.4cm}
\end{table}

\textbf{Comparison with State-of-the-arts.}
From the results in Table~\ref{aokvqa-sota}, it can be seen that in the multi-choice track, our two versions of AssistGPT (i.e., with light-weight BLIP2 FlanT5$_{XL}$ and more powerful Ins.BLIP Vicuna-7B) achieve the best among all current methods on in-context learning setting. It's worth noting that we use a pre-trained version of InstructBLIP, which performs at 53.3\%, as shown in Table~\ref{aokvqa-ablation}. When integrated into our system, it can enhance its performance to the level of fine-tuning model. For direct answer questions, while our performance on it may not match that of recently proposed models, it is still comparable to previous supervised SOTA, like GPV-2~\cite{kamath2022webly}. 

Our performance on direct answers did not surpass previous methods. The main reason is that for open-ended questions, models relying on LLM tend to output complete phrases rather than a single word as the final answer, even when we prompt them to provide as concise an answer as possible. For instance, for a given question "What flag is represented on the wall?", AssistGPT outputted the answer, "United States flag", but the correct answer does not include the word "flag", therefore it's deemed incorrect. This type of error is very common in AssistGPT. In the appendix, we show more examples to analyze the failure cases. Moreover, compared to the SOTA method, PromptCap~\cite{hu2022promptcap}, it specifically trained a caption model toward generating captions for A-OKVQA, which is also the reason for its good performance, while our system is more general.

From the results in Table~\ref{nextqa-sota}, AssistGPT achieved higher performance than recently proposed supervised methods, demonstrating the effectiveness of our approach. We can see that our model's performance mainly shows a more promising improvement in Causal and Descriptive questions, mainly due to our model continuously obtaining detailed information related to the question from the videos. Moreover, our method does not perform well on temporal questions. The main reason for this is that there are relatively few open-world temporal grounding models available, and mainstream work still involves fine-tuning on closed-world datasets. Therefore, we have to use the image captioner InstructBLIP with GPT-4 to achieve temporal grounding. The effect is not as good as that of fine-tuned models but has a more generalization ability. Furthermore, its performance is also very close to recent concurrent work, ViperGPT~\cite{suris2023vipergpt}. The ViperGPT is a little bit superior to ours, possibly because it has designed a sophisticated rule-based method, iteratively checking whether objects appear in the frame to perform temporal grounding.

\textbf{Ablation Study.}
We have designed several variants of AssistGPT to test the effectiveness of our proposed method. The most basic baseline is \textbf{InstructBLIP} (note that all following models are using Vicuna-7B version), which is the main source of visual information in AssistGPT. Since \textbf{InstructionBLIP} cannot necessarily output the answer in the required format, we design a variant, \textbf{InstructionBLIP+GPT-4} allows GPT-4 to further refine the output of InstructionBLIP. The \textbf{Reason-only} model directly plans all the steps the models need to run, similar to previous works~\cite{shen2023hugginggpt}. The \textbf{ReAct} model is capable of executing language-based ReAct. However, without Inspector and Code-like invocation forms, a subsequent model can only accept the output of the previous model, which is similar to ~\cite{yang2023mm}. We also ablate the Learner, which has three versions, \textbf{PIE} (i.e., w/o. Learner), \textbf{PEIL w. Self-Check} and \textbf{PEIL w. GT-Check}. 

From the results in Table~\ref{aokvqa-ablation}, we can see that the Reason-only model, which plans all the steps the models need to execute, showed a notable improvement in D.A. and M.C. This indicates that integrating multiple models can enhance model performance. The ReAct model, despite not having Inspector and Code-like invocation forms, showed a further improvement in both metrics, surpassing the Reason-only model. This suggests the effectiveness of ReAct manner. But involving our interleaved language and code, i.e., PIE, brings a more significant improvement on M.C.
Finally, the two variants of PIE with partial ablations, PEIL w. Self-Check and PEIL w. GT-Check, scored the highest on both tracks, showing the effectiveness of the Learner. The Learner shows a more significant improvement on D.A. tracks because models on D.A. often fail to output extremely short answers as required by A-OKVQA. The Learner can mitigate it by collecting in-context examples.

\begin{table}[]
\centering
\vspace{-0.4cm}
\caption{Comparison of our AssistGPT with SOTAs on NExT-QA dataset.}
\vspace{-0.1cm}
\label{nextqa-sota}
\resizebox{0.6\textwidth}{!}{%
\begin{tabular}{ll|ccc|c}
\toprule
                      & \multicolumn{1}{l|}{Method}                         & Causal & Temporal & Descriptive & All                       \\ \midrule
\multirow{5}{*}{\rotatebox{90}{\textcolor{gray!40}{Sup.}}} & \textcolor{gray!40}{HGA}                                                 & \textcolor{gray!40}{44.22}  & \textcolor{gray!40}{52.49}    & \textcolor{gray!40}{44.07}       & \textcolor{gray!40}{49.74}                     \\
                      & \textcolor{gray!40}{VQA-T~\cite{yang2021just}}     & \textcolor{gray!40}{49.60}  & \textcolor{gray!40}{51.49}    & \textcolor{gray!40}{63.19}       & \textcolor{gray!40}{52.32}                     \\
                      & \textcolor{gray!40}{ATP~\cite{buch2022revisiting}} & \textcolor{gray!40}{53.10}   & \textcolor{gray!40}{50.20}     & \textcolor{gray!40}{66.80}        & \textcolor{gray!40}{54.30}                      \\
                      & \textcolor{gray!40}{VGT~\cite{xiao2022video}}      & \textcolor{gray!40}{52.28}  & \textcolor{gray!40}{55.09}    & \textcolor{gray!40}{64.09}       & \textcolor{gray!40}{55.02}                     \\
                      & \textcolor{gray!40}{MIST~\cite{gao2022mist}}                                                & \textcolor{gray!40}{54.62}  & \textcolor{gray!40}{56.64}    & \textcolor{gray!40}{66.92}       & \textcolor{gray!40}{57.18}                     \\ \midrule
\multirow{2}{*}{\rotatebox{90}{ICL}}  & ViperGPT~\cite{suris2023vipergpt}                                            & -      & -        & -           & 60.00                      \\
                      & AssistGPT                                           & 60.02  & 51.38    & 67.26       & \multicolumn{1}{l}{58.36} \\ \bottomrule
\end{tabular}%
}
\end{table}

\subsection{Qualitative Results}
\begin{figure}[h]
  \centering
  \includegraphics[width=0.9\textwidth]{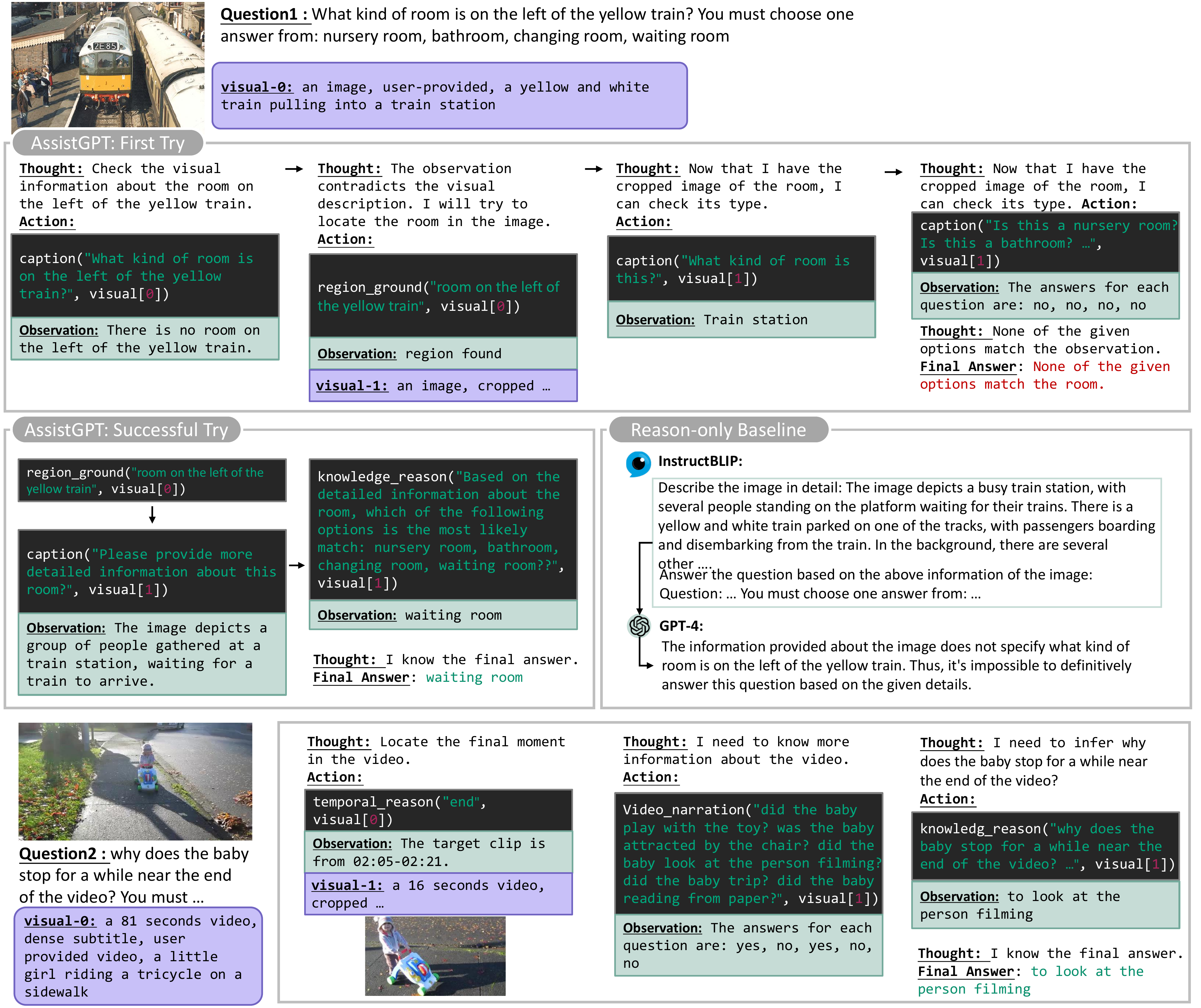}
  \caption{Qualitative results on A-OKVQA (Question 1) and NExT-QA dataset (Question 2).}
  \label{examples}
\end{figure}
In Fig.~\ref{examples}, we visualize some prediction cases from A-OKVQA (Question 1) and NExT-QA (Question 2). From both examples, it can be seen that AssistGPT can decompose the question into reasonable sub-tasks and then complete them step by step, ultimately obtaining the final answer. Moreover, due to the interleaved code and language reasoning method, the model can effectively invoke the necessary content as input. From the reasoning process of Question 1, we can also see AssistGPT's self-correction ability. When the visual model output unsatisfactory results, AssistGPT can dynamically invoke other modules, like the ground module to reason over another path. In addition, for Question 1, the model's first attempt did not yield effective results, and it will autonomously optimize the plan because it did not pass the self-check. In addition, we also present the result of the reason-only baseline. It first calls InstructBLIP to output a caption, then uses GPT-4 for inference. Since the information from the caption does not meet the requirements, resulting incorrect results. However, once the prediction fails, the model does not have a way to self-optimize. It's worth mentioning that the most significant feature of our method is that it can solve more complex problems than those in the benchmark, as the example in Fig.~\ref{fig1}. We show more in-the-wild examples in the Appendix.


\section{Conclusions and Limitations}
In this paper, we propose a novel multi-modal assistant system named AssistGPT that leverages an interleaved code and language reasoning approach, namely Plan, Execute, Inspect, and Learn (PEIL). This innovative system integrates LLM with various tools to address the challenges posed by complex visual-based tasks. Our experimental results on A-OKVQA and NExT-QA benchmarks demonstrate AssistGPT's effectiveness. Furthermore, we showcase our system's ability in handling diverse and intricate real-world scenarios. Our system also has some limitations. Our approach does not propose an end-to-end updating solution, which is crucial when the tools used make mistakes. Another limitation is that the planning process requires an extensive explanation of tools, resulting in a relatively large overhead, which could be improved by distilling a smaller size planner.


\renewcommand{\thesection}{\Alph{section}}
\setcounter{section}{0}


\section*{Appendix}
In the appendix, we provide additional details for the main paper:
\begin{itemize}[leftmargin=0.2in, topsep=0pt, itemsep=0pt]
    \item More discussion with existing modular systems in Sec.~\ref{secA}
    \item More details of AssistGPT in Sec.~\ref{secB}.
    \item More qualitative results of A-OKVQA in Sec.~\ref{secC}.
    \item More in-the-wild examples in Sec.~\ref{secD}.
\end{itemize}

\section{Discussion with Existing LLM-driven Modular Systems}
\label{secA}

\begin{table}[h]
\centering
\caption{\textbf{Comparison of existing LLM-driven modular systems.} We compare existing methods from four dimensions: Task Focus, Reasoning Method, Source Management (how they manage input and intermediate results), and whether they have learning capabilities. The term "ReAct" in the table does not strictly refer to using the ReAct~\cite{yao2022react}, but rather it denotes planning and executing concurrently.}
\label{comp}
\resizebox{\textwidth}{!}{%
\begin{tabular}{lcccccccc}
\toprule
\multicolumn{1}{c}{\multirow{2}{*}{Model}} & \multicolumn{3}{c}{Task Focus} & \multicolumn{2}{c}{Reasoning} & \multicolumn{2}{c}{Source Management} & \multirow{2}{*}{Learning} \\ \cmidrule(lr){2-4} \cmidrule(lr){5-6} \cmidrule(lr){7-8}
\multicolumn{1}{c}{}                       & NLP     & Image     & Video    & Format        & ReAct         & Input format    & \multicolumn{1}{c}{Method}    &                           \\ \midrule
Toolformer  \cite{schick2023toolformer}                               & \cmark  & \xmark    & \xmark   & lang. \& prog.      & \cmark        & text-only       & -                             &  \cmark                   \\
WebGPT \cite{nakano2021webgpt}                                    & \cmark  & \xmark    & \xmark   & program       & \cmark        & test-only       & -                             &  \cmark                   \\ \midrule
Visual ChatGPT \cite{wu2023visual}                            & \xmark  & \cmark    & \xmark   & language      & \xmark        & multi. V.  & Filename                      &  \xmark                   \\
ViperGPT  \cite{suris2023vipergpt}                                 & \xmark  & \cmark    & \cmark   & program       & \xmark        & single V.     & Variable                      &  \xmark                   \\
VisProg \cite{gupta2022visual}                                   & \xmark  & \cmark    & \xmark   & program       & \xmark        & single V.     & Variable                      &  \xmark                   \\
MM-ReAct \cite{yang2023mm}                                  &  \xmark  & \cmark    & \cxmark  & language      & \cmark        & multi V.   & Filename                      &  \xmark                   \\
Chameleon \cite{lu2023chameleon}                                 &  \cmark  & \cmark    & \xmark  & language      & \xmark        & single V.     & Cache update                  &  \xmark                   \\
HuggingGPT \cite{shen2023hugginggpt}                                &  \xmark & \cmark    & \cxmark   & language      & \xmark        & multi V.   & Filename                      &  \xmark                   \\ \midrule
AssistGPT (ours)                           &  \xmark  & \cmark    & \cmark  & lang. \& prog.  & \cmark      & multi V.   & Inspector                     &  \cmark                   \\ \bottomrule
\end{tabular}%
}
\end{table}

In Table~\ref{comp}, we compare the existing LLM-driven modular systems with our AssistGPT from four perspectives:

\textbf{Task Focus.} From the perspective of Task Focus, there are currently three works that can handle videos: Hugging GPT~\cite{shen2023hugginggpt}, MM-ReAct~\cite{yang2023mm}, and ViperGPT~\cite{suris2023vipergpt}. Hugging GPT and MM-ReAct merely demonstrate their capabilities in handling videos through a few simple examples (thus we mark them with \textcolor{BurntOrange}{orange checkmarks} \cxmark). For instance, Hugging GPT exhibits its video generation feature, while MM-ReAct showcases its ability to perform tasks such as summarization and localization based on subtitles. However, these methods have not been validated on any benchmark. ViperGPT can handle questions based on visual content. Compared to these works, AssistGPT is capable of dealing with more complex and general video question-answering tasks, including understanding subtitles, visual content, and OCR, and demonstrating long video comprehension capabilities.

\textbf{Reasoning.} In terms of reasoning, existing Multi-modal models primarily adopt a reason-only style, that is, directly deriving the solution steps based on the question. This approach struggles with handling complex visual inputs, and when the intermediate results don't meet expectations, the model also finds it hard to self-correct. MM-ReAct introduces the original ReAct for reasoning in Multi-modal tasks, but due to the original ReAct's inadequacy in dealing with complex non-text intermediate results, its current planning scheme for addressing video-related issues is basically two steps: extracting all information from the video and then having an LLM answer the question. 
In contrast, this paper proposes a more general Plan, Execute, Inspect, and Learn (PEIL) reasoning scheme. In the case of complex videos, our interleaved language and code reasoning approach allows for flexible language planning for the next step, and structured code for invoking input and intermediate results, thereby facilitating the handling of complex questions and visual content.

\textbf{Source Management.} Handling complex input and a large number of intermediate results is often crucial in complex reasoning processes. Current language-based reasoning methods mainly use filenames to label resources. Chameleon proposes an update mechanism with a cache that constantly updates the current reasoning results. Program-based reasoning, on the other hand, uses variables to store intermediate results. A deficiency of these methods is the inability of the language-based Planner to quickly comprehend the content of visual sources, which impedes the effective use of different sources to complete different subtasks. As a result, existing work struggles to handle flexible input and intermediate results. Even though some work supports multiple visual sources as input, they are more often batch-processed for similar tasks, with each source requiring similar operations. For instance, in HuggingGPT, the task of calculating the sum of the number of zebras in several images involves counting the number of zebras in each image. In contrast, our work introduces the Inspector, which records the metadata and summary of each visual source and provides it to the Planner for reasoning. This design can support complex input. For example, a user view image that describes the current user's problem, and a reference video as a source of knowledge, AssistGPT can then use these two different types of sources to jointly answer the user's question.

\textbf{Learning.} Most multi-modal modular systems lack the capability for continuous optimization. This paper proposes a simple update mechanism that allows the model to self-check the reasonableness of its output and ultimately continues to collect in-context learning examples.

\section{More details of AssistGPT}
\label{secB}

\begin{table}[h]
\centering
\caption{\textbf{Invoke Commands and Illustrations to the Modules in AssistGPT.}}
\label{tools}
\resizebox{\textwidth}{!}{%
\begin{tabular}{lll}
\toprule
\multicolumn{1}{c}{\textbf{Module}}         & \multicolumn{1}{c}{\textbf{Invoke Command}}                          & \multicolumn{1}{c}{\textbf{Illustration}}                                         \\ \midrule
(a) Image Caption    & {\scriptsize\colorbox{black}{\texttt{\textcolor{codey}{caption(}\textcolor{coder}{query}\textcolor{codeb}{, visual}\textcolor{codey}{[}\textcolor{codenum}{i}\textcolor{codey}{])}}}}    & extract the visual information in an image.                              \\
(b) Video Narration  & {\scriptsize\colorbox{black}{\texttt{\textcolor{codey}{video\_narration(}\textcolor{coder}{query}\textcolor{codeb}{, visual}\textcolor{codey}{[}\textcolor{codenum}{i}\textcolor{codey}{])}}}}  & output narration based on video's visual information.             \\
(c) Object Detection & {\scriptsize\colorbox{black}{\texttt{\textcolor{codey}{object\_detect(}\textcolor{coder}{query}\textcolor{codeb}{, visual}\textcolor{codey}{[}\textcolor{codenum}{i}\textcolor{codey}{])}}}} & detect required objects in an image.                                     \\
(d) Text Detection   & {\scriptsize\colorbox{black}{\texttt{\textcolor{codey}{text\_detect(}\textcolor{coder}{None}\textcolor{codeb}{, visual}\textcolor{codey}{[}\textcolor{codenum}{i}\textcolor{codey}{])}}}}    & extract the OCR in an image.                                             \\
(e) ASR Translation  & {\scriptsize\colorbox{black}{\texttt{\textcolor{codey}{asr(}\textcolor{coder}{None}\textcolor{codeb}{, visual}\textcolor{codey}{[}\textcolor{codenum}{i}\textcolor{codey}{])}}}}
               & transcribe audio to text.                                                \\
(f) Region Ground    & {\scriptsize\colorbox{black}{\texttt{\textcolor{codey}{region\_ground(}\textcolor{coder}{query}\textcolor{codeb}{, visual}\textcolor{codey}{[}\textcolor{codenum}{i}\textcolor{codey}{])}}}}     & locate the queried region in an image.                                   \\
(g) Narration Ground & {\scriptsize\colorbox{black}{\texttt{\textcolor{codey}{narration\_ground(}\textcolor{coder}{query}\textcolor{codeb}{, visual}\textcolor{codey}{[}\textcolor{codenum}{i}\textcolor{codey}{])}}}}  & find the clip based on the narration of a video.                         \\
(h) Text Ground      & {\scriptsize\colorbox{black}{\texttt{\textcolor{codey}{text\_ground(}\textcolor{coder}{query}\textcolor{codeb}{, visual}\textcolor{codey}{[}\textcolor{codenum}{i}\textcolor{codey}{])}}}}       & find the location of a specific text in an image.                        \\
(i) Subtitle Ground  & {\scriptsize\colorbox{black}{\texttt{\textcolor{codey}{subtitle\_ground(}\textcolor{coder}{query}\textcolor{codeb}{, visual}\textcolor{codey}{[}\textcolor{codenum}{i}\textcolor{codey}{])}}}}   & find the clip based on the subtitle of a video.                          \\
(j) Knowledge Reason & {\scriptsize\colorbox{black}{\texttt{\textcolor{codey}{knowledge\_reason(}\textcolor{coder}{query}\textcolor{codeb}{, }\textcolor{codey}{[}\textcolor{codenum}{}\textcolor{codey}{])}}}}        & infer the answer based on the commonsense.                 \\
(k) Narration Reason & {\scriptsize\colorbox{black}{\texttt{\textcolor{codey}{narration\_reason(}\textcolor{coder}{query}\textcolor{codeb}{, visual}\textcolor{codey}{[}\textcolor{codenum}{i}\textcolor{codey}{])}}}}  & infer the answer based on narration of a video.                          \\
(l) Subtitle Reason  & {\scriptsize\colorbox{black}{\texttt{\textcolor{codey}{subtitle\_reason(}\textcolor{coder}{query}\textcolor{codeb}{, visual}\textcolor{codey}{[}\textcolor{codenum}{i}\textcolor{codey}{])}}}}  & infer the answer based on subtitle of a video.                           \\
(m) Temporal Reason  & {\scriptsize\colorbox{black}{\texttt{\textcolor{codey}{temporal\_reason(}\textcolor{coder}{query}\textcolor{codeb}{, visual}\textcolor{codey}{[}\textcolor{codenum}{i}\textcolor{codey}{])}}}}  & find the clip based on temporal relationship words. \\ \bottomrule
\end{tabular}%
}
\end{table}

In Table~\ref{tools}, we show the invoke commands and illustration of each module in AssistGPT. We provide more details of how each module is implemented.

\begin{itemize}[leftmargin=0.2in, topsep=0pt, itemsep=1pt]
    \item \textbf{Image Caption}: The core model of this module is a text-conditioned captioning model, e.g., BLIP2~\cite{li2023blip}, InstructBLIP~\cite{dai2023instructblip}, similar to an open-ended Visual Question Answering model.
    \item \textbf{Video Narration}: As the general video captioning models are not yet mature, we currently use the image captioning model~\cite{li2023blip, dai2023instructblip} to accomplish this function. Specifically, we sample image frames (1/3 FPS for current implementation) and perform text-conditioned captioning on each frame. We employ text-conditioned captioning because, if we use dense captioning, the output text will be excessively abundant, making it difficult for subsequent models to utilize. The Video Narration feature can also optionally read the OCR content within the frames. The extracted OCR will be appended to the caption of each frame.
    \item \textbf{Object Detection}: The main function of this module is to determine whether the image contains the objects mentioned in the query and to address counting-related questions. Thus, it contains an open-set object detection model, e.g., Grounding DINO~\cite{liu2023grounding}, which can output the bounding boxes of relevant objects based on the query. We also let the module calculate the number of related objects. 
    \item \textbf{Text Detection}: This model is used to extract OCR from images, and the extracted text is returned to the Planner. We use Google OCR to achieve this purpose.
    \item \textbf{ASR Translation}: This model is used to convert audio from a video into text. We use OpenAI's open-source ASR (Automatic Speech Recognition) model, Whisper~\cite{radford2022robust}, to accomplish this. The detected ASR organizes timestamps and text in a manner similar to subtitles. In the implementation, we automatically run this module as soon as we receive a video with audio.
    \item \textbf{Region Ground}: The purpose of this module is to find a specific area of an image based on the query. We use the OFA-Large~\cite{wang2022ofa}, which is fine-tuned on RefCOCO, to achieve it.
    \item \textbf{Narration Ground}: This model's function is to find time segments related to the query based on the video's narration. We propose two implementations: 1) We use GPT-4~\cite{openai2022chatgpt}, taking the video's narration and query as prompts, to output the timestamps of the time segments. 2) Another solution is using CLIP~\cite{radford2021learning} to do that. We can split the video into several segments, and calculate the similarity between the frame in each segment and query. The time stamps of the segment with the highest similarity will be outputted. In our preliminary experiments, the first solution showed better interpretability and generalization ability, so it was adopted in the benchmark evaluation.
    \item \textbf{Text Ground}: The purpose of this model is to locate specific areas of an image that correspond to a certain text. This capability can guide users in identifying crucial information in complex, text-rich images, such as user interfaces. The query format is \texttt{text[:object\_name]}, wherein \texttt{text} signifies the text to be located, and \texttt{object\_name} (which is optional) is used to locate the text on a specific object, for instance, "menu: button".
    Specifically, the model operates in two stages: 
    1) Based on the Optical Character Recognition (OCR) detection results, the model identifies areas of the image that match the text segment of the query. This is achieved by calculating the distance between the query and the OCR extracted, and when the edit distance is below a particular threshold, it is considered a match. 
    2) If more than one textual area is identified, we further refine the results based on the object's name. We employ the Semantic Segment Anything (SSA)~\cite{chen2023semantic} to segment the image semantically, identifying regions that match the object's name mentioned in the query.
    \item \textbf{Subtitle Ground}: This model is similar to the narration grounding model, but it uses the video's subtitles as input instead of the narration. Thus, we also use GPT-4 to achieve it.
    \item \textbf{Knowledge Reason}: The purpose of this model is to enable the model to apply external knowledge to answer questions. We currently do not connect to the internet to retrieve knowledge, but use the knowledge that GPT-4 has itself learned. Specifically, this model enables GPT-4 to use its own knowledge to infer the answer based on the question and results of all previous reasoning steps.
    \item \textbf{Narration Reason}: The aim of this module is to infer some information based on the visual content of the video. This module also uses GPT-4, taking the query and the input video's narration as prompts, to infer the answer.
    \item \textbf{Subtitle Reason}: The aim of this module is to infer some information based on the subtitle of the video. It is similar to Narration Reason, but takes the input video's subtitle and query as prompts, to infer the answer.
    \item \textbf{Temporal Reason}: This model is able to find a video clip based on some temporal relation words. The input to this module follows the following format: temporal\_word: time stamps, e.g., \texttt{after: 3 - 6}. Temporal relation words include two types, one is absolute temporal relation words, such as in the middle/beginning/end of the video. The second type is relative temporal relation words, such as before and after. For the first type of words, we divide the video into 5 segments and then output the time stamps of the corresponding segment according to the temporal\_word. For the second type, we divide the video into 8 segments, and then, according to the input time stamps, we output the time stamps of the segment before or after it. The current hyperparameters, the division of video clips, are still preliminary. It would be much better to use the model to divide them semantically, and then perform temporal reasoning in the future.
\end{itemize}

\section{Qualitative Results in A-OKVQA}
\label{secC}
\begin{figure}[ht]
  \includegraphics[width=\textwidth]{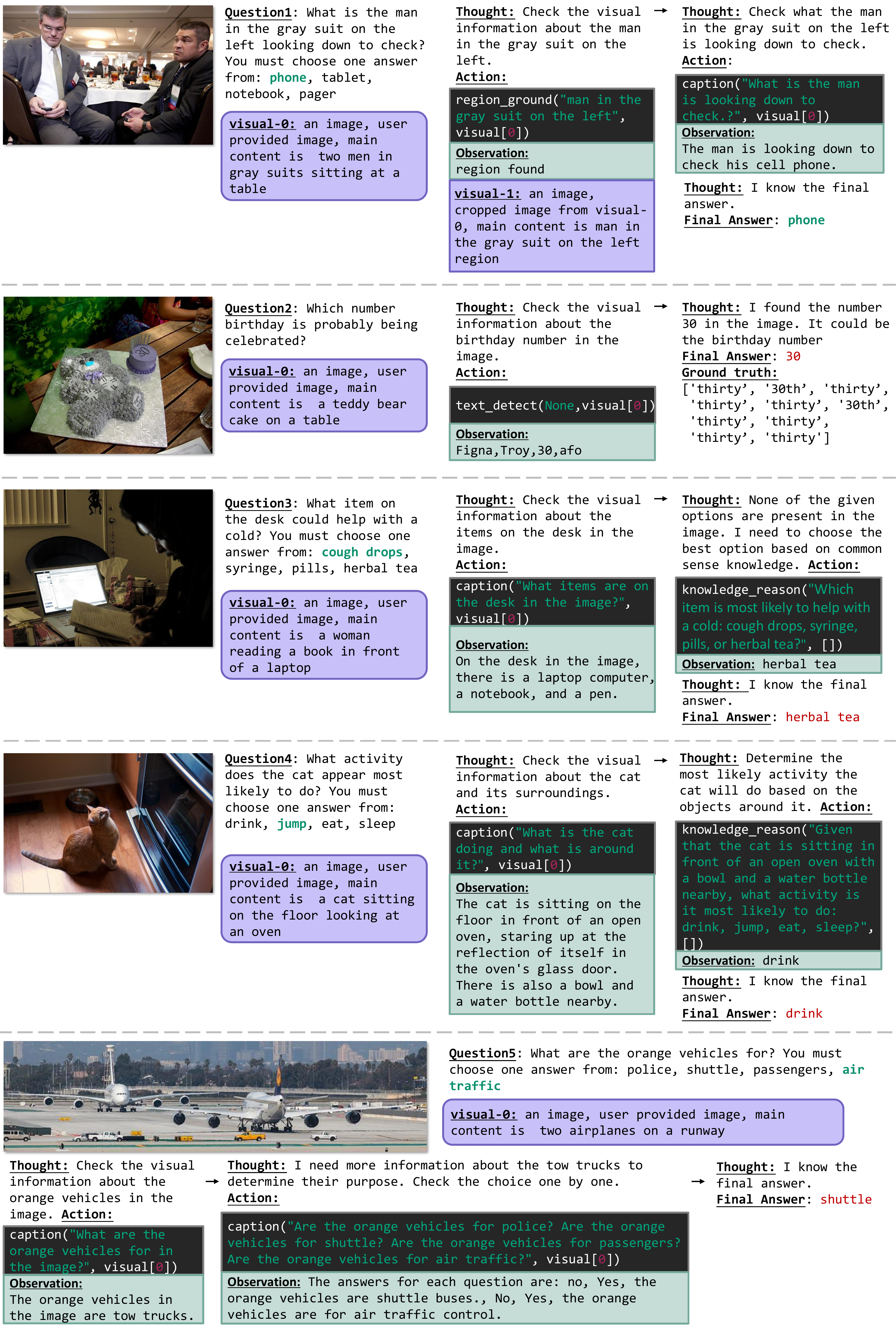}
  \caption{\textbf{Reasoning process of AssistGPT on A-OKVQA.} The choice colored with \textbf{\textcolor{mygreen}{green}} in the question indicates the ground truth.}
  \label{sup_aokvqa_example}
\end{figure}

In Figure~\ref{sup_aokvqa_example}, we showcase a successful instance along with several failure examples, illustrating the most frequent error patterns in A-OKVQA. As is evident, AssistGPT can produce highly interpretable answer processes. Moreover, even in cases where the questions are answered incorrectly, there are relatively reasonable explanations provided. In the following, we illustrate the common error patterns in detail:
\begin{itemize}[leftmargin=0.2in, topsep=0pt, itemsep=1pt]
    \item \textbf{Undesired output format}: For Direct Answer questions, like Q2, the results of AssistGPT are the same as the correct answers in meaning, but the expression is different, which would be considered as incorrect under the existing metrics.
    \item \textbf{Fine-grained recognition}: The recognition of fine-grained categories of some objects is still not well done by existing visual models, resulting in the incorrect final answer. For example, AssistGPT didn't successfully recognize cough drops in Q3. 
    \item \textbf{Pose-to-text}: Currently, there are very few models that can map the fine-grained pose or actions of people or animals to natural language. For example, capturing the upward jump action of the cat in Q4 is a challenge. AssistGPT currently does not incorporate a related model to grasp such information. Instead, it makes prediction based on the surrounding objects in relation to the cat.
    \item \textbf{Inconsistent reasoning}: Despite AssistGPT having some self-error correction mechanisms, it occasionally exhibits inconsistencies in its reasoning process, which can lead to final inaccuracies. For instance, in Q5, the model initially identifies the orange vehicle as a truck, but in subsequent steps, it is referred to as a shuttle bus. Unfortunately, AssistGPT fails to detect this inconsistency and does not proceed to make necessary corrections.
    
\end{itemize}

\section{In-the-wild Prediction Examples}
\label{secD}
\begin{figure}[ht]
  \includegraphics[width=\textwidth]{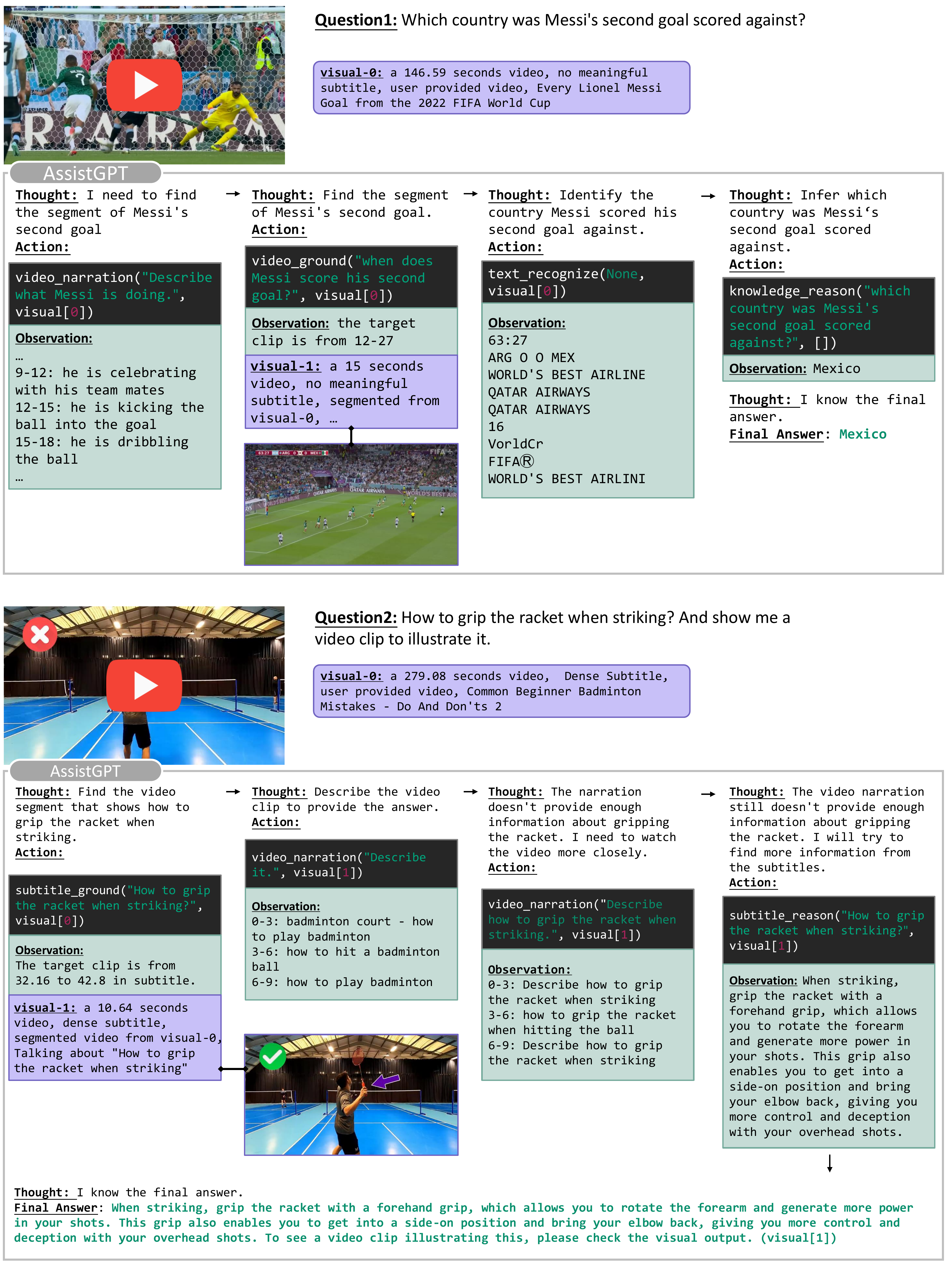}
  \caption{\textbf{The reasoning process of AssistGPT when handling in-the-wild questions.}}
  \label{sup_in_the_wild_example1}
\end{figure}

\begin{figure}[!ht]
  \includegraphics[width=\textwidth]{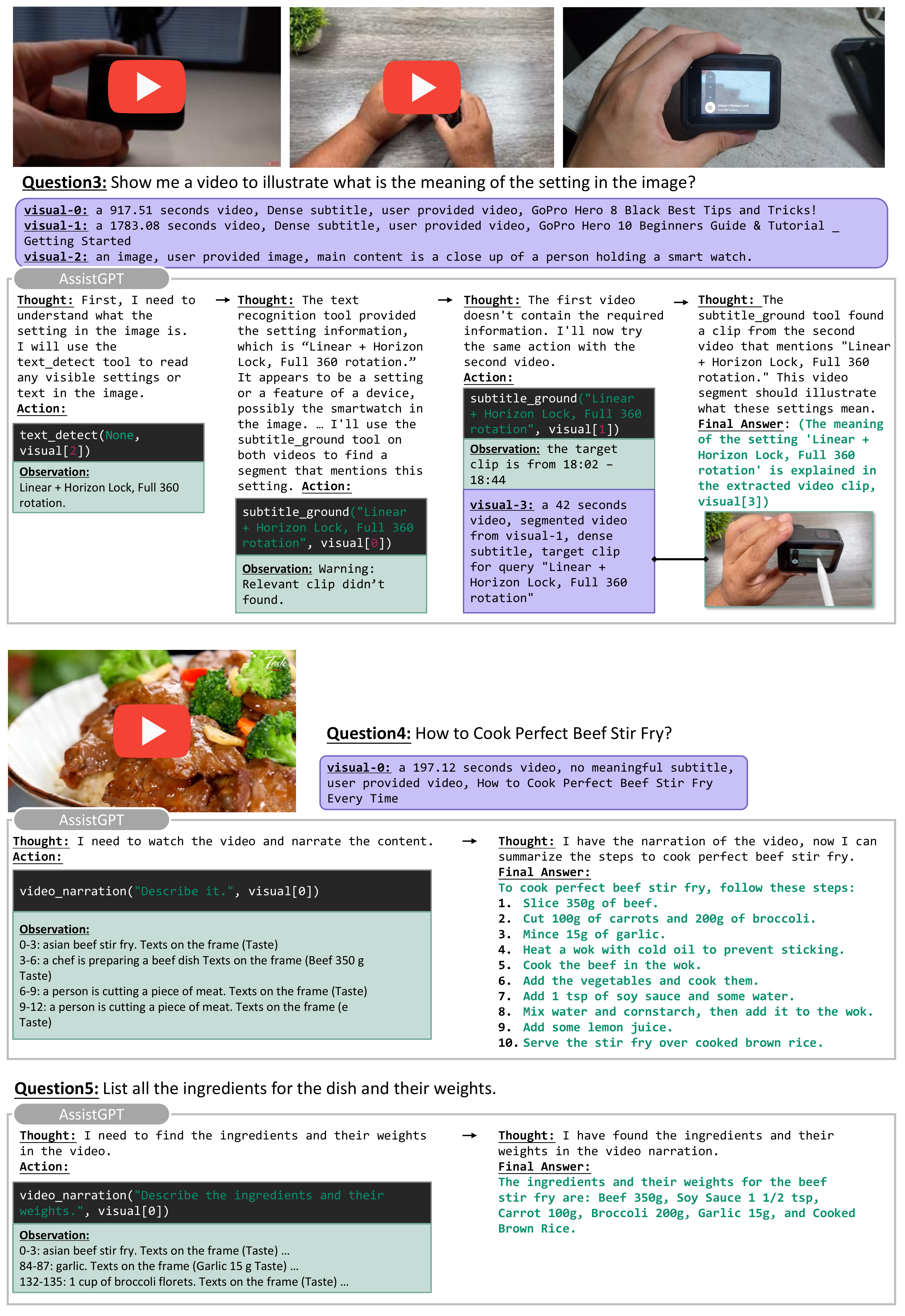}
  \caption{\textbf{The reasoning process of AssistGPT when handling in-the-wild questions.}}
  \label{sup_in_the_wild_example2}
\end{figure}

We show some examples of AssistGPT handling in-the-wild scenarios in Figure~\ref{sup_in_the_wild_example1} and Figure~\ref{sup_in_the_wild_example2}. From various in-the-wild examples, it's clear that AssistGPT can adeptly handle a range of video types, be it dense, subtitled instructional videos (Q2, Q3), or those featuring rich, visual content with sporadic on-frame text (Q1, Q4, Q5). Impressively, when faced with high-level queries (Q2 and Q3), the model exhibits a capacity to strategically locate useful content, accurately identify the correct responses, and offer comprehensive, multimodal answers. A notable self-error correction capability is also evident during its reasoning process, as demonstrated in Q2. Here, the narration model was unable to generate meaningful narrations and, therefore, opted to utilize the subtitle to answer the question.

Moreover, in Q5, we highlight that our model can effectively process multiple video inputs serving different functions. This includes a User view image and a couple of reference videos. It's important to note that our model can accommodate any number of inputs. Consequently, with the incorporation of a YouTube video search function, the model could autonomously seek out several reference videos and then cross-reference them to discern the user's intent.

In summary, we want to emphasize that AssistGPT is a comprehensive multi-modal assistant system, capable of managing a wide array of real-world application queries that are far more complex and comprehensive than the samples provided in benchmarks.

\clearpage

\bibliographystyle{unsrt}
\bibliography{ref}

\begin{thebibliography}{10}

\bibitem{bert}
Jacob Devlin, Ming-Wei Chang, Kenton Lee, and Kristina Toutanova.
\newblock Bert: Pre-training of deep bidirectional transformers for language
  understanding.
\newblock {\em arXiv preprint arXiv:1810.04805}, 2018.

\bibitem{gpt2}
Alec Radford, Jeffrey Wu, Rewon Child, David Luan, Dario Amodei, Ilya
  Sutskever, et~al.
\newblock Language models are unsupervised multitask learners.
\newblock {\em OpenAI blog}, page~9, 2019.

\bibitem{gpt3}
Tom Brown, Benjamin Mann, Nick Ryder, Melanie Subbiah, Jared~D Kaplan, Prafulla
  Dhariwal, Arvind Neelakantan, Pranav Shyam, Girish Sastry, Amanda Askell,
  et~al.
\newblock Language models are few-shot learners.
\newblock {\em NeurIPS}, pages 1877--1901, 2020.

\bibitem{llama}
Hugo Touvron, Thibaut Lavril, Gautier Izacard, Xavier Martinet, Marie-Anne
  Lachaux, Timoth{\'e}e Lacroix, Baptiste Rozi{\`e}re, Naman Goyal, Eric
  Hambro, Faisal Azhar, et~al.
\newblock Llama: Open and efficient foundation language models.
\newblock {\em arXiv preprint arXiv:2302.13971}, 2023.

\bibitem{openai2022chatgpt}
OpenAI.
\newblock Introducing chatgpt.
\newblock OpenAI Blog, 09 2021.

\bibitem{zeng2022socratic}
Andy Zeng, Adrian Wong, Stefan Welker, Krzysztof Choromanski, Federico Tombari,
  Aveek Purohit, Michael Ryoo, Vikas Sindhwani, Johnny Lee, Vincent Vanhoucke,
  et~al.
\newblock Socratic models: Composing zero-shot multimodal reasoning with
  language.
\newblock {\em arXiv preprint arXiv:2204.00598}, 2022.

\bibitem{wang2022language}
Zhenhailong Wang, Manling Li, Ruochen Xu, Luowei Zhou, Jie Lei, Xudong Lin,
  Shuohang Wang, Ziyi Yang, Chenguang Zhu, Derek Hoiem, et~al.
\newblock Language models with image descriptors are strong few-shot
  video-language learners.
\newblock {\em arXiv preprint arXiv:2205.10747}, 2022.

\bibitem{wu2023visual}
Chenfei Wu, Shengming Yin, Weizhen Qi, Xiaodong Wang, Zecheng Tang, and Nan
  Duan.
\newblock Visual chatgpt: Talking, drawing and editing with visual foundation
  models.
\newblock {\em arXiv preprint arXiv:2303.04671}, 2023.

\bibitem{blip}
Junnan Li, Dongxu Li, Caiming Xiong, and Steven Hoi.
\newblock Blip: Bootstrapping language-image pre-training for unified
  vision-language understanding and generation.
\newblock In {\em ICML}, pages 12888--12900, 2022.

\bibitem{blip2}
Junnan Li, Dongxu Li, Silvio Savarese, and Steven Hoi.
\newblock Blip-2: Bootstrapping language-image pre-training with frozen image
  encoders and large language models.
\newblock {\em arXiv preprint arXiv:2301.12597}, 2023.

\bibitem{dai2023instructblip}
Wenliang Dai, Junnan Li, Dongxu Li, Anthony Meng~Huat Tiong, Junqi Zhao,
  Weisheng Wang, Boyang Li, Pascale Fung, and Steven Hoi.
\newblock Instructblip: Towards general-purpose vision-language models with
  instruction tuning.
\newblock {\em arXiv preprint arXiv:2305.06500}, 2023.

\bibitem{detr}
Nicolas Carion, Francisco Massa, Gabriel Synnaeve, Nicolas Usunier, Alexander
  Kirillov, and Sergey Zagoruyko.
\newblock End-to-end object detection with transformers.
\newblock In {\em ECCV}, pages 213--229, 2020.

\bibitem{Li_2022_CVPR}
Liunian~Harold Li, Pengchuan Zhang, Haotian Zhang, Jianwei Yang, Chunyuan Li,
  Yiwu Zhong, Lijuan Wang, Lu~Yuan, Lei Zhang, Jenq-Neng Hwang, Kai-Wei Chang,
  and Jianfeng Gao.
\newblock Grounded language-image pre-training.
\newblock In {\em Proceedings of the IEEE/CVF Conference on Computer Vision and
  Pattern Recognition (CVPR)}, pages 10965--10975, June 2022.

\bibitem{wu2022grit}
Jialian Wu, Jianfeng Wang, Zhengyuan Yang, Zhe Gan, Zicheng Liu, Junsong Yuan,
  and Lijuan Wang.
\newblock Grit: A generative region-to-text transformer for object
  understanding.
\newblock {\em arXiv preprint arXiv:2212.00280}, 2022.

\bibitem{wang2022git}
Jianfeng Wang, Zhengyuan Yang, Xiaowei Hu, Linjie Li, Kevin Lin, Zhe Gan,
  Zicheng Liu, Ce~Liu, and Lijuan Wang.
\newblock Git: A generative image-to-text transformer for vision and language.
\newblock {\em arXiv preprint arXiv:2205.14100}, 2022.

\bibitem{chen2022pali}
Xi~Chen, Xiao Wang, Soravit Changpinyo, AJ~Piergiovanni, Piotr Padlewski,
  Daniel Salz, Sebastian Goodman, Adam Grycner, Basil Mustafa, Lucas Beyer,
  Alexander Kolesnikov, Joan Puigcerver, Nan Ding, Keran Rong, Hassan Akbari,
  Gaurav Mishra, Linting Xue, Ashish Thapliyal, James Bradbury, Weicheng Kuo,
  Mojtaba Seyedhosseini, Chao Jia, Burcu~Karagol Ayan, Carlos Riquelme, Andreas
  Steiner, Anelia Angelova, Xiaohua Zhai, Neil Houlsby, and Radu Soricut.
\newblock Pali: A jointly-scaled multilingual language-image model, 2022.

\bibitem{shen2023hugginggpt}
Yongliang Shen, Kaitao Song, Xu~Tan, Dongsheng Li, Weiming Lu, and Yueting
  Zhuang.
\newblock Hugginggpt: Solving ai tasks with chatgpt and its friends in
  huggingface.
\newblock {\em arXiv preprint arXiv:2303.17580}, 2023.

\bibitem{ge2023openagi}
Yingqiang Ge, Wenyue Hua, Jianchao Ji, Juntao Tan, Shuyuan Xu, and Yongfeng
  Zhang.
\newblock Openagi: When llm meets domain experts.
\newblock {\em arXiv preprint arXiv:2304.04370}, 2023.

\bibitem{lu2023chameleon}
Pan Lu, Baolin Peng, Hao Cheng, Michel Galley, Kai-Wei Chang, Ying~Nian Wu,
  Song-Chun Zhu, and Jianfeng Gao.
\newblock Chameleon: Plug-and-play compositional reasoning with large language
  models.
\newblock {\em arXiv preprint arXiv:2304.09842}, 2023.

\bibitem{li2023videochat}
KunChang Li, Yinan He, Yi~Wang, Yizhuo Li, Wenhai Wang, Ping Luo, Yali Wang,
  Limin Wang, and Yu~Qiao.
\newblock Videochat: Chat-centric video understanding.
\newblock {\em arXiv preprint arXiv:2305.06355}, 2023.

\bibitem{suris2023vipergpt}
D{\'\i}dac Sur{\'\i}s, Sachit Menon, and Carl Vondrick.
\newblock Vipergpt: Visual inference via python execution for reasoning.
\newblock {\em arXiv preprint arXiv:2303.08128}, 2023.

\bibitem{gupta2022visual}
Tanmay Gupta and Aniruddha Kembhavi.
\newblock Visual programming: Compositional visual reasoning without training.
\newblock {\em arXiv preprint arXiv:2211.11559}, 2022.

\bibitem{duan2023taskmatrix}
Yaobo Liang, Chenfei Wu, Ting Song, Wenshan Wu, Yan Xia, Yu~Liu, Yang Ou, Shuai
  Lu, Lei Ji, Shaoguang Mao, Yun Wang, Linjun Shou, Ming Gong, and Nan Duan.
\newblock Taskmatrix.ai: Completing tasks by connecting foundation models with
  millions of apis.
\newblock {\em arXiv preprint arXiv:2303.16434}, 2023.

\bibitem{yao2022react}
Shunyu Yao, Jeffrey Zhao, Dian Yu, Nan Du, Izhak Shafran, Karthik Narasimhan,
  and Yuan Cao.
\newblock React: Synergizing reasoning and acting in language models.
\newblock {\em arXiv preprint arXiv:2210.03629}, 2022.

\bibitem{schick2023toolformer}
Timo Schick, Jane Dwivedi-Yu, Roberto Dess{\`\i}, Roberta Raileanu, Maria
  Lomeli, Luke Zettlemoyer, Nicola Cancedda, and Thomas Scialom.
\newblock Toolformer: Language models can teach themselves to use tools.
\newblock {\em arXiv preprint arXiv:2302.04761}, 2023.

\bibitem{antol2015vqa}
Stanislaw Antol, Aishwarya Agrawal, Jiasen Lu, Margaret Mitchell, Dhruv Batra,
  C~Lawrence Zitnick, and Devi Parikh.
\newblock Vqa: Visual question answering.
\newblock In {\em Proceedings of the IEEE international conference on computer
  vision}, pages 2425--2433, 2015.

\bibitem{anderson2018bottom}
Peter Anderson, Xiaodong He, Chris Buehler, Damien Teney, Mark Johnson, Stephen
  Gould, and Lei Zhang.
\newblock Bottom-up and top-down attention for image captioning and visual
  question answering.
\newblock In {\em Proceedings of the IEEE conference on computer vision and
  pattern recognition}, pages 6077--6086, 2018.

\bibitem{lu2019vilbert}
Jiasen Lu, Dhruv Batra, Devi Parikh, and Stefan Lee.
\newblock Vilbert: Pretraining task-agnostic visiolinguistic representations
  for vision-and-language tasks.
\newblock {\em Advances in neural information processing systems}, 32, 2019.

\bibitem{tan2019lxmert}
Hao Tan and Mohit Bansal.
\newblock Lxmert: Learning cross-modality encoder representations from
  transformers.
\newblock {\em arXiv preprint arXiv:1908.07490}, 2019.

\bibitem{gao2020learning}
Difei Gao, Ruiping Wang, Shiguang Shan, and Xilin Chen.
\newblock Learning to recognize visual concepts for visual question answering
  with structural label space.
\newblock {\em IEEE Journal of Selected Topics in Signal Processing},
  14(3):494--505, 2020.

\bibitem{gao2017visual}
Difei Gao, Ruiping Wang, Shiguang Shan, and Xilin Chen.
\newblock Visual textbook network: Watch carefully before answering visual
  questions.
\newblock In {\em BMVC}, 2017.

\bibitem{marino2019ok}
Kenneth Marino, Mohammad Rastegari, Ali Farhadi, and Roozbeh Mottaghi.
\newblock Ok-vqa: A visual question answering benchmark requiring external
  knowledge.
\newblock In {\em Proceedings of the IEEE/cvf conference on computer vision and
  pattern recognition}, pages 3195--3204, 2019.

\bibitem{aokvqa}
Dustin Schwenk, Apoorv Khandelwal, Christopher Clark, Kenneth Marino, and
  Roozbeh Mottaghi.
\newblock A-okvqa: A benchmark for visual question answering using world
  knowledge.
\newblock {\em arXiv}, 2022.

\bibitem{wang2017fvqa}
Peng Wang, Qi~Wu, Chunhua Shen, Anthony Dick, and Anton Van Den~Hengel.
\newblock Fvqa: Fact-based visual question answering.
\newblock {\em IEEE transactions on pattern analysis and machine intelligence},
  40(10):2413--2427, 2017.

\bibitem{gui2021kat}
Liangke Gui, Borui Wang, Qiuyuan Huang, Alex Hauptmann, Yonatan Bisk, and
  Jianfeng Gao.
\newblock Kat: A knowledge augmented transformer for vision-and-language.
\newblock {\em arXiv preprint arXiv:2112.08614}, 2021.

\bibitem{marino2021krisp}
Kenneth Marino, Xinlei Chen, Devi Parikh, Abhinav Gupta, and Marcus Rohrbach.
\newblock Krisp: Integrating implicit and symbolic knowledge for open-domain
  knowledge-based vqa.
\newblock In {\em Proceedings of the IEEE/CVF Conference on Computer Vision and
  Pattern Recognition}, pages 14111--14121, 2021.

\bibitem{gao2022cric}
Difei Gao, Ruiping Wang, Shiguang Shan, and Xilin Chen.
\newblock Cric: A vqa dataset for compositional reasoning on vision and
  commonsense.
\newblock {\em IEEE Transactions on Pattern Analysis and Machine Intelligence},
  2022.

\bibitem{bain2021frozen}
Max Bain, Arsha Nagrani, G{\"u}l Varol, and Andrew Zisserman.
\newblock Frozen in time: A joint video and image encoder for end-to-end
  retrieval.
\newblock In {\em Proceedings of the IEEE/CVF International Conference on
  Computer Vision}, pages 1728--1738, 2021.

\bibitem{Lei_2021_CVPR}
Jie Lei, Linjie Li, Luowei Zhou, Zhe Gan, Tamara~L. Berg, Mohit Bansal, and
  Jingjing Liu.
\newblock Less is more: Clipbert for video-and-language learning via sparse
  sampling.
\newblock In {\em Proceedings of the IEEE/CVF Conference on Computer Vision and
  Pattern Recognition (CVPR)}, pages 7331--7341, June 2021.

\bibitem{wang2022geb+}
Yuxuan Wang, Difei Gao, Licheng Yu, Weixian Lei, Matt Feiszli, and Mike~Zheng
  Shou.
\newblock Geb+: A benchmark for generic event boundary captioning, grounding
  and retrieval.
\newblock In {\em Computer Vision--ECCV 2022: 17th European Conference, Tel
  Aviv, Israel, October 23--27, 2022, Proceedings, Part XXXV}, pages 709--725.
  Springer, 2022.

\bibitem{grauman2022ego4d}
Kristen Grauman, Andrew Westbury, Eugene Byrne, Zachary Chavis, Antonino
  Furnari, Rohit Girdhar, Jackson Hamburger, Hao Jiang, Miao Liu, Xingyu Liu,
  et~al.
\newblock Ego4d: Around the world in 3,000 hours of egocentric video.
\newblock In {\em Proceedings of the IEEE/CVF Conference on Computer Vision and
  Pattern Recognition}, pages 18995--19012, 2022.

\bibitem{lin2022egocentric}
Kevin~Qinghong Lin, Jinpeng Wang, Mattia Soldan, Michael Wray, Rui Yan, Eric~Z
  XU, Difei Gao, Rong-Cheng Tu, Wenzhe Zhao, Weijie Kong, et~al.
\newblock Egocentric video-language pretraining.
\newblock {\em Advances in Neural Information Processing Systems},
  35:7575--7586, 2022.

\bibitem{gao2021env}
Difei Gao, Ruiping Wang, Ziyi Bai, and Xilin Chen.
\newblock Env-qa: A video question answering benchmark for comprehensive
  understanding of dynamic environments.
\newblock In {\em Proceedings of the IEEE/CVF International Conference on
  Computer Vision}, pages 1675--1685, 2021.

\bibitem{hou2022cone}
Zhijian Hou, Wanjun Zhong, Lei Ji, Difei Gao, Kun Yan, Wing-Kwong Chan,
  Chong-Wah Ngo, Zheng Shou, and Nan Duan.
\newblock Cone: An efficient coarse-to-fine alignment framework for long video
  temporal grounding.
\newblock {\em arXiv preprint arXiv:2209.10918}, 2022.

\bibitem{wong2022assistq}
Benita Wong, Joya Chen, You Wu, Stan~Weixian Lei, Dongxing Mao, Difei Gao, and
  Mike~Zheng Shou.
\newblock Assistq: Affordance-centric question-driven task completion for
  egocentric assistant.
\newblock In {\em European Conference on Computer Vision}, pages 485--501.
  Springer, 2022.

\bibitem{lei2022assistsr}
Weixian Lei, Difei Gao, Yuxuan Wang, Dongxing Mao, Zihan Liang, Lingmin Ran,
  and Mike~Zheng Shou.
\newblock Assistsr: Task-oriented video segment retrieval for personal ai
  assistant.
\newblock In {\em Findings of the Association for Computational Linguistics:
  EMNLP 2022}, pages 319--338, 2022.

\bibitem{chen2023affordance}
Joya Chen, Difei Gao, Kevin~Qinghong Lin, and Mike~Zheng Shou.
\newblock Affordance grounding from demonstration video to target image.
\newblock In {\em Proceedings of the IEEE/CVF Conference on Computer Vision and
  Pattern Recognition}, pages 6799--6808, 2023.

\bibitem{hu2020iterative}
Ronghang Hu, Amanpreet Singh, Trevor Darrell, and Marcus Rohrbach.
\newblock Iterative answer prediction with pointer-augmented multimodal
  transformers for textvqa.
\newblock In {\em Proceedings of the IEEE/CVF Conference on Computer Vision and
  Pattern Recognition}, pages 9992--10002, 2020.

\bibitem{yang2021tap}
Zhengyuan Yang, Yijuan Lu, Jianfeng Wang, Xi~Yin, Dinei Florencio, Lijuan Wang,
  Cha Zhang, Lei Zhang, and Jiebo Luo.
\newblock Tap: Text-aware pre-training for text-vqa and text-caption.
\newblock In {\em Proceedings of the IEEE/CVF conference on computer vision and
  pattern recognition}, pages 8751--8761, 2021.

\bibitem{Gao_2020_CVPR}
Difei Gao, Ke~Li, Ruiping Wang, Shiguang Shan, and Xilin Chen.
\newblock Multi-modal graph neural network for joint reasoning on vision and
  scene text.
\newblock In {\em Proceedings of the IEEE/CVF Conference on Computer Vision and
  Pattern Recognition (CVPR)}, June 2020.

\bibitem{lei2022symbolic}
Stan~Weixian Lei, Difei Gao, Jay~Zhangjie Wu, Yuxuan Wang, Wei Liu, Mengmi
  Zhang, and Mike~Zheng Shou.
\newblock Symbolic replay: Scene graph as prompt for continual learning on vqa
  task.
\newblock {\em arXiv preprint arXiv:2208.12037}, 2022.

\bibitem{openai2023gpt4}
OpenAI.
\newblock Gpt-4 technical report, 2023.

\bibitem{driess2023palm}
Danny Driess, Fei Xia, Mehdi~SM Sajjadi, Corey Lynch, Aakanksha Chowdhery,
  Brian Ichter, Ayzaan Wahid, Jonathan Tompson, Quan Vuong, Tianhe Yu, et~al.
\newblock Palm-e: An embodied multimodal language model.
\newblock {\em arXiv preprint arXiv:2303.03378}, 2023.

\bibitem{li2023blip}
Junnan Li, Dongxu Li, Silvio Savarese, and Steven Hoi.
\newblock Blip-2: Bootstrapping language-image pre-training with frozen image
  encoders and large language models.
\newblock {\em arXiv preprint arXiv:2301.12597}, 2023.

\bibitem{zhu2022minigpt4}
Deyao Zhu, Jun Chen, Xiaoqian Shen, xiang Li, and Mohamed Elhoseiny.
\newblock Minigpt-4: Enhancing vision-language understanding with advanced
  large language models, 2023.

\bibitem{chen2021evaluating}
Mark Chen, Jerry Tworek, Heewoo Jun, Qiming Yuan, Henrique Ponde de~Oliveira
  Pinto, Jared Kaplan, Harri Edwards, Yuri Burda, Nicholas Joseph, Greg
  Brockman, et~al.
\newblock Evaluating large language models trained on code.
\newblock {\em arXiv preprint arXiv:2107.03374}, 2021.

\bibitem{andreas2016neural}
Jacob Andreas, Marcus Rohrbach, Trevor Darrell, and Dan Klein.
\newblock Neural module networks.
\newblock In {\em Proceedings of the IEEE conference on computer vision and
  pattern recognition}, pages 39--48, 2016.

\bibitem{hu2017learning}
Ronghang Hu, Jacob Andreas, Marcus Rohrbach, Trevor Darrell, and Kate Saenko.
\newblock Learning to reason: End-to-end module networks for visual question
  answering.
\newblock In {\em Proceedings of the IEEE international conference on computer
  vision}, pages 804--813, 2017.

\bibitem{johnson2017inferring}
Justin Johnson, Bharath Hariharan, Laurens Van Der~Maaten, Judy Hoffman,
  Li~Fei-Fei, C~Lawrence~Zitnick, and Ross Girshick.
\newblock Inferring and executing programs for visual reasoning.
\newblock In {\em Proceedings of the IEEE international conference on computer
  vision}, pages 2989--2998, 2017.

\bibitem{wei2022chain}
Jason Wei, Xuezhi Wang, Dale Schuurmans, Maarten Bosma, Ed~Chi, Quoc Le, and
  Denny Zhou.
\newblock Chain of thought prompting elicits reasoning in large language
  models.
\newblock {\em arXiv preprint arXiv:2201.11903}, 2022.

\bibitem{yang2023mm}
Zhengyuan Yang, Linjie Li, Jianfeng Wang, Kevin Lin, Ehsan Azarnasab, Faisal
  Ahmed, Zicheng Liu, Ce~Liu, Michael Zeng, and Lijuan Wang.
\newblock Mm-react: Prompting chatgpt for multimodal reasoning and action.
\newblock {\em arXiv preprint arXiv:2303.11381}, 2023.

\bibitem{kirillov2023segment}
Alexander Kirillov, Eric Mintun, Nikhila Ravi, Hanzi Mao, Chloe Rolland, Laura
  Gustafson, Tete Xiao, Spencer Whitehead, Alexander~C Berg, Wan-Yen Lo, et~al.
\newblock Segment anything.
\newblock {\em arXiv preprint arXiv:2304.02643}, 2023.

\bibitem{chen2023semantic}
Jiaqi Chen, Zeyu Yang, and Li~Zhang.
\newblock Semantic segment anything.
\newblock \url{https://github.com/fudan-zvg/Semantic-Segment-Anything}, 2023.

\bibitem{liu2023grounding}
Shilong Liu, Zhaoyang Zeng, Tianhe Ren, Feng Li, Hao Zhang, Jie Yang, Chunyuan
  Li, Jianwei Yang, Hang Su, Jun Zhu, et~al.
\newblock Grounding dino: Marrying dino with grounded pre-training for open-set
  object detection.
\newblock {\em arXiv preprint arXiv:2303.05499}, 2023.

\bibitem{radford2022robust}
Alec Radford, Jong~Wook Kim, Tao Xu, Greg Brockman, Christine McLeavey, and
  Ilya Sutskever.
\newblock Robust speech recognition via large-scale weak supervision.
\newblock {\em arXiv preprint arXiv:2212.04356}, 2022.

\bibitem{wang2022ofa}
Peng Wang, An~Yang, Rui Men, Junyang Lin, Shuai Bai, Zhikang Li, Jianxin Ma,
  Chang Zhou, Jingren Zhou, and Hongxia Yang.
\newblock Ofa: Unifying architectures, tasks, and modalities through a simple
  sequence-to-sequence learning framework.
\newblock {\em CoRR}, abs/2202.03052, 2022.

\bibitem{radford2021learning}
Alec Radford, Jong~Wook Kim, Chris Hallacy, Aditya Ramesh, Gabriel Goh,
  Sandhini Agarwal, Girish Sastry, Amanda Askell, Pamela Mishkin, Jack Clark,
  et~al.
\newblock Learning transferable visual models from natural language
  supervision.
\newblock In {\em International Conference on Machine Learning}, pages
  8748--8763. PMLR, 2021.

\bibitem{xiao2021next}
Junbin Xiao, Xindi Shang, Angela Yao, and Tat-Seng Chua.
\newblock Next-qa: Next phase of question-answering to explaining temporal
  actions.
\newblock In {\em Proceedings of the IEEE/CVF Conference on Computer Vision and
  Pattern Recognition}, pages 9777--9786, 2021.

\bibitem{kamath2022webly}
Amita Kamath, Christopher Clark, Tanmay Gupta, Eric Kolve, Derek Hoiem, and
  Aniruddha Kembhavi.
\newblock Webly supervised concept expansion for general purpose vision models.
\newblock In {\em Computer Vision--ECCV 2022: 17th European Conference, Tel
  Aviv, Israel, October 23--27, 2022, Proceedings, Part XXXVI}, pages 662--681.
  Springer, 2022.

\bibitem{hu2022promptcap}
Yushi Hu, Hang Hua, Zhengyuan Yang, Weijia Shi, Noah~A Smith, and Jiebo Luo.
\newblock Promptcap: Prompt-guided task-aware image captioning.
\newblock {\em arXiv preprint arXiv:2211.09699}, 2022.

\bibitem{yang2021just}
Antoine Yang, Antoine Miech, Josef Sivic, Ivan Laptev, and Cordelia Schmid.
\newblock Just ask: Learning to answer questions from millions of narrated
  videos.
\newblock In {\em Proceedings of the IEEE/CVF International Conference on
  Computer Vision}, pages 1686--1697, 2021.

\bibitem{buch2022revisiting}
Shyamal Buch, Crist{\'o}bal Eyzaguirre, Adrien Gaidon, Jiajun Wu, Li~Fei-Fei,
  and Juan~Carlos Niebles.
\newblock Revisiting the" video" in video-language understanding.
\newblock In {\em Proceedings of the IEEE/CVF Conference on Computer Vision and
  Pattern Recognition}, pages 2917--2927, 2022.

\bibitem{xiao2022video}
Junbin Xiao, Pan Zhou, Tat-Seng Chua, and Shuicheng Yan.
\newblock Video graph transformer for video question answering.
\newblock In {\em European Conference on Computer Vision}, pages 39--58.
  Springer, 2022.

\bibitem{gao2022mist}
Difei Gao, Luowei Zhou, Lei Ji, Linchao Zhu, Yi~Yang, and Mike~Zheng Shou.
\newblock Mist: Multi-modal iterative spatial-temporal transformer for
  long-form video question answering.
\newblock {\em arXiv preprint arXiv:2212.09522}, 2022.

\bibitem{nakano2021webgpt}
Reiichiro Nakano, Jacob Hilton, Suchir Balaji, Jeff Wu, Long Ouyang, Christina
  Kim, Christopher Hesse, Shantanu Jain, Vineet Kosaraju, William Saunders,
  et~al.
\newblock Webgpt: Browser-assisted question-answering with human feedback.
\newblock {\em arXiv preprint arXiv:2112.09332}, 2021.

\end{thebibliography}

\end{document}